\documentclass{article}

\usepackage{arxiv}

\usepackage[utf8]{inputenc} % allow utf-8 input
\usepackage[T1]{fontenc}    % use 8-bit T1 fonts
\usepackage{hyperref}       % hyperlinks
\usepackage{url}            % simple URL typesetting
\usepackage{booktabs}       % professional-quality tables
\usepackage{amsfonts}       % blackboard math symbols
\usepackage{nicefrac}       % compact symbols for 1/2, etc.
\usepackage{microtype}      % microtypography
\usepackage{lipsum}

\usepackage{graphicx}
\usepackage{amsmath}

\title{Yet it moves: Learning from Generic Motions to Generate IMU data from YouTube videos.}

\author {
    Vitor Fortes Rey \\
    DFKI \\
    Kaiserslautern, Germany \\
    \texttt{vitor.fortes@dfki.de} \\
    \And
    Kamalveer Kaur Garewal \\
    former DFKI \\
    Kaiserslautern, Germany \\
    \texttt{kamalveer.garewal@gmail.com} \\
    \And
    Paul Lukowicz \\
    DFKI, University of Kaiserslautern \\
    Kaiserslautern, Germany \\
    \texttt{paul.lukowicz@dfki.de} \\
}

\begin{document}
\maketitle

\begin{abstract}
Human activity recognition (HAR) using wearable sensors has benefited much less from recent advances in Machine Learning than fields such as computer vision and natural language processing.  This is to a large extent due to the lack of large scale repositories of  labeled training data. In our research we aim to  facilitate the use of online videos, which exists in ample quantity for most activities and are much easier to label than sensor data, to  simulate labeled wearable motion sensor data. In previous work we already demonstrate some preliminary results in this direction focusing on very simple, activity specific simulation models and a single sensor modality (acceleration norm)\cite{10.1145/3341162.3345590}.  In this paper we show how we can train a regression model on generic motions for both accelerometer and gyro signals and then apply it to videos of the target activities to generate synthetic IMU data (acceleration and gyro norms) that can be used to train and/or improve HAR models.  We demonstrate that systems trained on simulated data generated by our  regression model  can come to within around 10\% of the mean F1 score of a system trained on real sensor data. Furthermore we show that by either including a small amount of real sensor data for model calibration or simply leveraging the fact that (in general) we can easily generate  much more simulated data from video than we can collect in terms of real sensor data the advantage of real sensor data can be eventually equalized.
\end{abstract}

% keywords can be removed

\section{Introduction and Related Work}
Human activity recognition (HAR) using wearable sensors has been a successful research field \cite{lara2012survey,wang2019deep}. As  wearable sensors are becoming ubiquitous complex HAR systems are moving out of the lab into long term, real life scenarios. There has been the expectation that as large amounts of sensor data become available HAR will be able to benefit from the advances in Deep Learning techniques in the same way as fields such as computer vision and speech recognition. However, while the application of deep learning methods to HAR has produced undeniable advances \cite{wang2019deep} the progress has so far been  far less significant  then than computer vision or NLP. Thus for example in recent HAR competitions most deep learning entries are out-performed by classical machine learning methods, namely random forests\cite{10.1145/3410530.3414341}, while in computer vision deep learning techniques have dominated since 2012\cite{ILSVRC15} and beyond\cite{knoll2020advancing}. 

It is widely believed that the above is in a large degree due the difficulty of acquiring   {\bf labeled} sensor data  from complex realistic scenarios.   Today there are not only billions of images and videos  freely available online, but they can be quickly and easily labeled using crowdsourcing services such as Amazon Mechanical Turk, by leveraging captions, or known topics of image collections. By contrast while the availability of sensor data as such is rapidly increasing with the spread of various sensor enabled personal devices,  labeling such data remains a challenge. Whereas most image labelling tasks can be easily accomplished by a very  broad public (anyone can say if an image shows a cat or a dog), interpreting raw sensor data can be challenging even for experts. As a consequence using crowdsourcing for labeling large amounts of sensor data is not a viable option. Neither is having users provide labels for their own data, at least not over long time periods and on a large scale.  Even if many users are willing to provide sensor data from their smart watches, smart phones, smart shoes, smart rockets etc (at least as long as privacy concerns are properly addressed), only very few are willing to make the effort needed to annotate such data, especially not on a fine grained basis needed for many HAR problems (min by min) and not over weeks or months (which is what reaching the amount of training data comparable to computer vision would require).  
\subsection{Paper Contributions}
Recently the idea of using labeled videos to generate "synthetic" sensor data has been proposed as a solution for the HAR labeled data problem. As one of the first published approaches in this direction we have previously shown \cite{10.1145/3341162.3345590}  initial results on how regression models can be trained to simulate motion sensor data from videos of the respective activities.  In that preliminary work we  relied on recordings  that contained sensor data and videos \emph{of the same activities} that we later wanted to have the system generate the sensor data for.   
In this paper we describe the full  developed and evaluation that  initial  idea making the following contributions:
\begin{enumerate}
    \item We have adapted the  method to generalise outside the target exercises. This  means that our model can now be trained  on other subjects performing a set of "generic" motions selected to be representative of a broad domain.  Using this
generic regression model, we can generate synthetic  training  data for a variety of different activities.
\item We have performed and described an in depth analysis of the physical background of deriving different components of the  IMU (inertial measurement unit) signal and have shown how to simulate different sensor signals beyond acceleration norm (as given in the initial paper).
\item We have performed an  analysis of  the influence of various signal processing techniques on the quality of the simulated sensor signal.
\item We have proposed and implemented a deep neural network based regression model for the generation of simulated sensor data from videos.
\item We  have identified a set of generic motions suitable for training the regression model for the broad domain of aerobics like physical exercises, recorded a dataset containing appropriate video footage and sensor data and used it to train the above model. 
\item We have performed elaborate evaluation on an activity recognition task from the above broad domain of aerobics like exercises for which we have both recorded our own dataset and collected and used as source of simulated training data appropriate online videos.  This includes the analysis of the influence of simulating different sensor modalities and adding small amounts of real sensor data to the simulated one to further improve accuracy. 
Compared to our previous preliminary work \cite{10.1145/3341162.3345590}, we achieve significantly better  better results,  although the problem that we tackle (using {\bf generic motions} for training the regression model) is a harder one. Overall we show that 
\begin{enumerate}
    \item Systems trained on simulated data generated by our regression model and tested on real sensor data can achieve a recognition performance that is within 10\% of the recognition performance achieved by a model trained on the corresponding real sensor data.
    \item By increasing the amount of simulated data we can match the performance of a real signal based model trained on less data. Results so far indicate about a factor of 2 to be needed. This is in line with the motivation behind this work which is the fact that very large amount of videos are available for many relevant activities online which in turn means that we can get orders of magnitude more training data if we can generate it from such videos (which we aim to enable).
    \item Adding even small amounts of real sensor data to fine tune the model generated on the basis of simulated data can improve the performance significantly. In our experiments we found that real  data from only 1 or 2 real users can already make the performance of a system trained in simulated data comparable to one fully trained on real data.
\end{enumerate}

\end{enumerate}
\begin{figure}
    \centering
    \includegraphics[width=1.0\columnwidth]{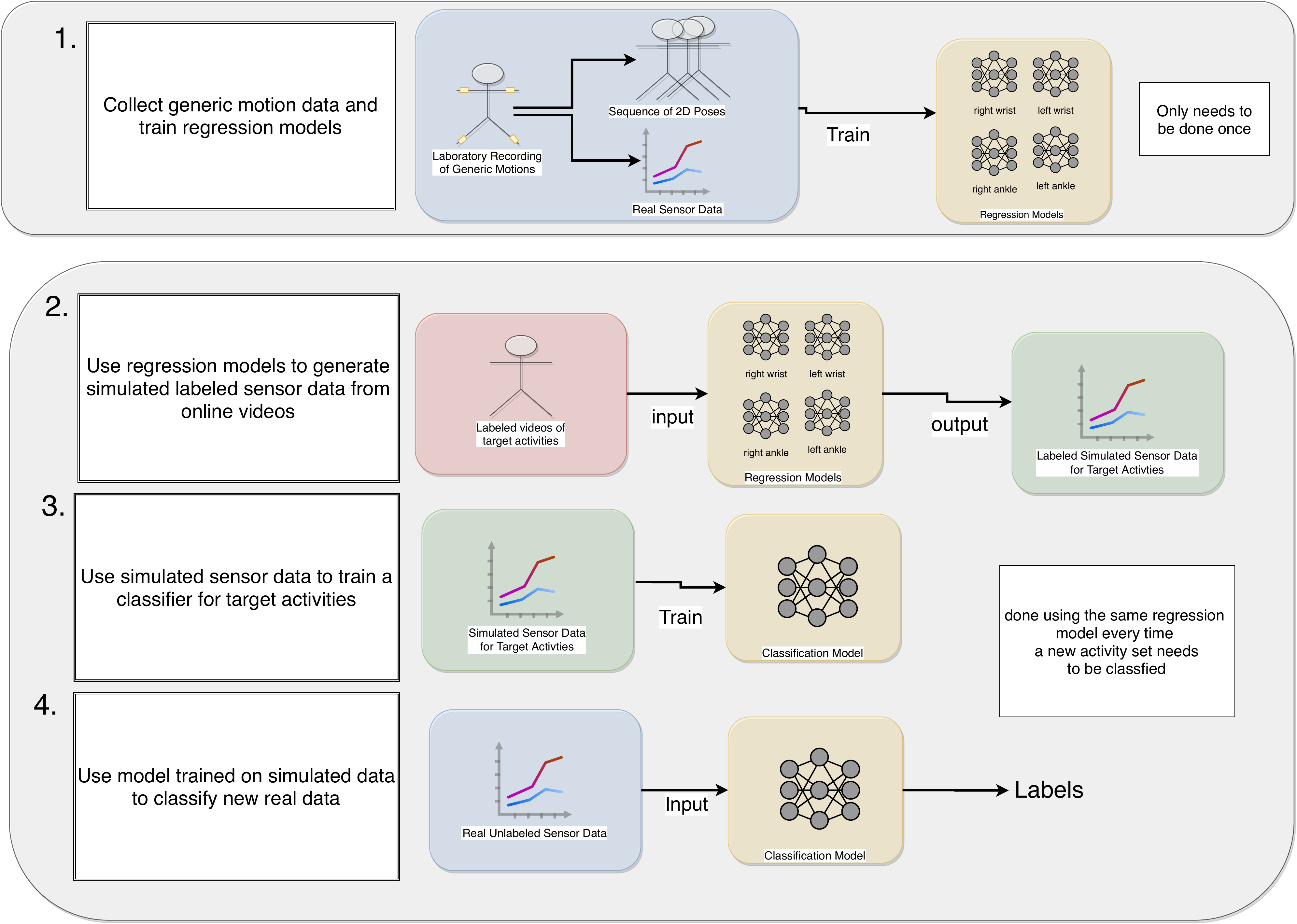}
    \caption{The steps involved in our overall method.}
    \label{fig:step}
\end{figure}
The resulting overall process that we propose to use  videos for training sensor based activity recognition systems can be seen in Figure \ref{fig:step}.
First (step 1) we record a dataset that contains a broad set of generic motions typical for a given domain (eg.physical exercise). This dataset contains both an appropriate video recording and the corresponding data from on body sensors. We then train a regression model to map the videos onto the corresponding sensor data. The development of this model is the main contribution of our work. 

Note that a regression model needs to be generated only once for a given broad domain. Once it is trained it can  be repeatedly used to generate simulated sensor data for various arbitrary activity recognition tasks from that respective domain. To this, for each new recognition task a set of labeled videos of the respective a activities is first acquired (eg. from YouTube). The videos are then fed to the regression model (step 2) which converts it to simulated sensor data. Obviously the labels for the sensor data are identical with the video labels. Thus we have labeled training sensor data. We next use it to train a recognition model for the required activities (step 3). Finally the trained model is used to recognize those activities from real
sensor data in the end application (step 4).
 
 \subsection{Related Work}
Learning using data from simulated environments is not novel in itself, as it is common in fields such as robotics and others.
In the field of HAR, related work regarding obtaining this simulated data for HAR can be separated in the following way: 
\begin{enumerate}
    \item Methods that work directly with simulated environments and through them obtain sensor values. For example, if one has a 3D model of a person walking in some physical simulator, one can obtain the acceleration measured by an IMU in that person's wrist. While plenty of physical simulators exist, creating realistic animations for all possible activities or even many of the different ways people perform the complex ones is cumbersome.
    \item Methods that aim to create a realistic 3D animation model from a real recording, which was obtained with, for example, one or more cameras, and using this model then to obtain the sensor values in a physical simulator.
    Those approaches can in theory take advantage of data from real subjects obtained from the many video repositories available, instead of relying directly on 3D modelling. On the other hand, those methods work best when two or more cameras are recording a scene, which is not the standard. Moreover, even for methods that work on monocular cameras, the signals obtained are only as good as the obtained 3D trajectories. While impressive results have been obtained recently for 3D pose estimation using monocular cameras, it remains a non-trivial problem.
    \item Methods that rely real on real recordings, but use machine learning approaches to simulate the sensor data directly instead of obtaining it from the physical simulation. As the previous one, this approach has the advantage of not needing to directly do any 3D modeling of the activities, with the difference that the machine learning task shifts from obtaining reliable 3D poses to directly obtaining sensor signals from the available information. Our method is in this category as we do not create a 3D representation of poses, but learn to map sequences of 2D poses to sequences of sensor values.
\end{enumerate}
Thus, we will first cover related work in the building blocks of IMU simulation and at last discuss how our method relates to current approaches.
\subsection{Simulating IMU Data Directly from Virtual Environments}
Many methods exist for simulating IMU data. For example, \cite{asare2013bodysim, ascher2010new, young2011imusim} provides a simulation environment where one can model human subjects with virtual sensor models to allow experimentation and exploration in virtual space of scenarios in inertial sensing. Depending on the application, specialised tools can be used. For example, there is ample work in simulating foot IMU data for pedestrian dead reckoning\cite{zampella2011simulation}. In general many tools exist in this field as it is useful for many projects\cite{smith2003simulation,pares2008yet}, but one may also use any existing physical simulator.

\subsection{Obtaining 3D Poses from Videos}
Recently, pose estimation has been a very interesting and successful area of research, with many applications.
Obtaining those 3D poses is possible with a calibrated system of more than one camera\cite{cao2018openpose}.
Commercial systems that can extract poses using $2$ cameras (among other things) includes also kinect, which was used in other work\cite{6246149} to obtain the 3D poses and with it simulate IMU signals. 

Using a single camera is obviously more challenging. Works in this are include \cite{kanazawa2018end, elhayek2018fully,mehta2017vnect,rogez2019lcr,onorina2017,omran2018neural}. 
Some works \cite{elhayek2018fully,onorina2017} predict the 2D joints and then fit the 3D skeleton using different strategies, others \cite{kanazawa2018end,rogez2019lcr,mehta2017vnect} are trained to predict 3D joints directly from the image. Even more interesting, some methods predict both 2D and 3D keypoints\cite{omran2018neural}. While methods have achieved impressive results, the field is still evolving and the task itself is far from easy due to occlusion, clothing, lighting, and the inherent ambiguity in inferring 3D from 2D \cite{bogo2016keep}.

Regarding notable current approaches for simulating HAR sensor data, we have \cite{asare2013bodysim} directly in simulated environments. Other notable approaches rely on body markers\cite{Takeda:2018}\cite{lago2019measured} or systems with multiple cameras such as kinect\cite{6246149}. Our approach does not use any special markers, but relies directly on the body joint's positions as we would not have access to such markers in video sources such as YouTube. Moreover, we do not assume we have more than a single monocular camera, as that is the case for most datasets and video sources. 
Regarding works where sensor data is simulated using forward kinematics, our approach does not directly perform physical simulation, as it is only feasible if the 3D poses of the subject are estimated. Obtaining those 3D poses is possible with a calibrated system of more than one camera\cite{cao2018openpose} or even using a single camera\cite{kanazawa2018end, elhayek2018fully,mehta2017vnect,rogez2019lcr,onorina2017,omran2018neural}. 
Thus, one could extract those poses and then use the cited methods to obtain the sensor values, but their quality depends on the quality of poses that can be obtained from a monocular camera. 
A recent work that follows this approach is\cite{kwon2020imutube}, which uses many off the shelf components to predict 3D human poses for monocular images and then forward kinematics to simulate sensor data. They have demonstrated the feasibility of the approach in generating acceleration data without a static camera, but their pipeline works best when predicting modes of locomotion and not complex activities that do not happen in a 2D plane\cite{kwon2020imutube}, probably due to accumulated distortions in the predicted 3D poses. 
Extracting those poses from a monocular camera is no easy task due to occlusion, clothing, lighting, and the inherent ambiguity in inferring 3D from 2D \cite{bogo2016keep}.
Our approach is different as we predict sensor data directly from the 2D poses using a deep neural network without forward kinematics or a complex pipeline that includes several deep neural networks.

We situate our work in the area of applying advances in vision to the HAR field. More recently vision has been shown to be successful for both labeling and knowledge transfer in mobile sensor based HAR \cite{10.1145/3351242}, showing that there are many benefits in using video information too for HAR and many cases where it can aid in data acquisition.
\section{Problem Description and Design Considerations}
\label{sec:problem}
On the surface the problem may seem fairly simple. A video (at least when taken from the right perspective) clearly contains elaborate motion information. In fact when watching a video of an activity a human gets such a  good idea how the subjects in the video move that very often videos are used to teach people complex motions (e.g. in sports). On the other  hand signals produced by the sensors that we are considering (IMUs and their components: accelerometers and gyroscope) are  in effect a representation of motion. As described in Related Work the generation of IMU signals from  6DOF (x,y,z coordinates plus the yaw, roll pitch angles) trajectories is a well understood and solved task. This in simplified terms we consider the mapping from one representation of relevant motion (video) to another (IMU sensors).
The difficulty that we need to address stems from three considerations:
\begin{enumerate}
    \item {\bf Fundamental incompleteness of 2D video information.} Monocular video (which is what we need to work with to be able to harvest online video sources) does not contain complete 6-DOF information on objects (in our case human body parts) that it shows. Instead for each object a single video frame provides 2D coordinates in its own frame of reference which  are essentially the x- and y-coordinates in pixels.  When looking at such a frame humans can translate such 2D image coordinates into  information about physical coordinates (which in some cases may include all 6 DOFs) through semantic analysis of the picture and putting the results of such semantic analysis in the context of their understanding of the world. Thus when seeing a man waving his arms we can estimate the physical coordinates of various body parts with respect to each other just from our knowledge of human physiology (degrees of freedom of joints, typical motions typical size and proportions of human body etc.). Clearly not knowing the exactly dimensions of the specific person this can only be an approximation.  If the person is for example a goal keeper standing on the goal line  
    the knowledge of (1) the dimensions of the goal an (2) the knowledge of the fact that the goal line is vertically underneath the goal post allows us an even better estimation including an estimate of the z-coordinate), however it remains an estimate, not an exact value.    
    \item {\bf Inherent inaccuracies in 2D video information.} Even when physical coordinates of relevant body parts can be in principle inferred from the semantic information, the achievable accuracy can greatly vary depending on the camera angle  and position. Thus, for example,  given frontal view of the user (as for example in Figure \ref{fig:ex_ex} the x-y position  of the wrist (point 4 in  wrist Figure \ref{fig:model_separation}) in respect to the hip (point 8) can be estimated with reasonable accuracy. On the other hand, the angle of rotation of the wrist around the lower arm axis is  much more difficult to estimate accurately. This is because  in a typical video  the wrist itself can be just  a few pixels wide.  Thus  large changes in terms of a rotation angle may correspond to change in one or two pixels only  significantly limiting  the available angular resolution.  
    \item  {\bf  Sensors specific physical effects.} Sensor signals are influenced by factors which do not directly manifest themselves in a image. Best example is gravity. While the direction of "down" can mostly be inferred from semantic analysis of an image it is not always enough. Consider the example of the wrist rotation above in the context of a wrist worn acceleration sensor. If the lower arm is parallel to the ground then the rotation determines how the gravity vector is projected onto the sensor axis perpendicular to the lower arm. Thus, the rotation has a very significant influence on the value  of the acceleration signals on those two axis. On the other hand, as described above, wrist rotation tends to produce a "weak signal" in an image so that it can only be coarsely estimated.  Another example is high frequency "ringing" of the acceleration signals when for example the user stamps his feet on the ground. It results from high frequency vibration of the device caused by the strong impact of the foot on a hard surface. It is a very characteristic feature in the acceleration signals, but mostly invisible on the video signal as the amplitude of the vibrations is too small and  effects the device only (not body parts as a whole) which may be invisible (e,g. hidden under clothing).
    \item{\bf Frame of reference and scaling.} IMUs typically use "world" coordinate system which relies on magnetic sensors to determine geographic north  and derives the direction of the gravity vector as down. The signals are given in standard units (e.g. $\frac{m}{s^2}$ for acceleration). 
    As mentioned above gravity can be in principle be derived through semantic analysis of an image, which is however not necessarily trivial and always reliable. Obviously in most images "north" is not given.  All motions are given in pixels per frame with the relationship between pixels and physical units depending on the camera position and settings.
    Overall the translation between image frame of reference and pixel units and the sensor frame of reference and physical units is s a non trivial challenge. 
\end{enumerate}
From the above discussion it is clear that the generation of IMU sensor data from videos cannot be accomplished with an exact physical model. This is a fundamental difference to the well understood and largely solved problem of generating IMU values from 6DOF trajectories. Instead a heuristic approach is needed that minimizes the unavoidable errors with respect to the needs of a given  domain. Here a key question is whether we narrowly optimize the system to a specific domain or follow a more general domain agnostic approach. Other core design concerns are how much, at which stage in the process and in what form  can be accomplished with exact physical models and which part of the task will be delegated to what type of data driven ML approaches. The most obvious choice is between (1) using existing ML methods for estimating 3D poses from 2D videos \cite{kanazawa2018end, elhayek2018fully,mehta2017vnect,rogez2019lcr,onorina2017} and then using bio mechanical models \cite{asare2013bodysim} to generate the sensor data and (2) training an end-to-end ML model that goes directly from 2D video to sensor data. This includes the question of, on the video side, how to accomplish the conversion between image frame of reference and pixel coordinates and, on the sensor's side, how to handle the sensor frame of reference and its physical units and whether filtering, scaling, and signal normalization should be done explicitly or left to the ML models.

\section{Method}\label{sec_method}
\begin{figure}
    \centering
    \includegraphics[width=1.0\columnwidth]{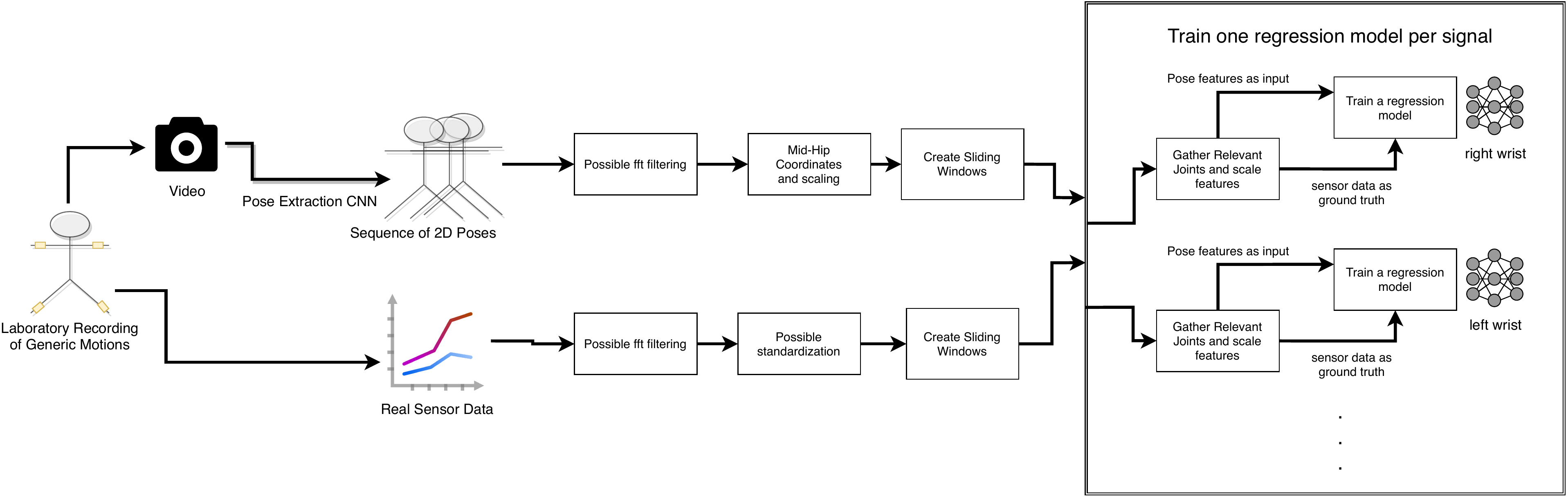}
    \caption{Procedure for training our regression models, which translates sequences of poses to the sensor values they generate. Using a dataset that includes both video and sensor data, we train for each position and sensor channel a regression model that uses the relevant joints and other features to predict the sensor values for that channel and place. }
    \label{fig:train}
\end{figure}
As described in the Introduction the core of our contribution is the development of a  model  for  generating simulated sensor data from videos of the respective activities.   The architecture of the model is shown in Figure \ref{fig:train}.    It consists of three stages: 
\begin{enumerate}
    \item The application of standard tools (specifically Open Pose \cite{cao2018openpose,cao2017realtime}) to identify humans in videos and extractions of their {\bf 2D} poses. The poses are defined in terms of joint coordinates in pixels.
    \item  A sliding window based translation of the raw  2D pixel coordinates into appropriately filtered and scaled values in the body centered (hip as origin) frame of reference.  This window considers $1.5$ seconds in the past and $1.5$ in the future, with a total length of $3$ seconds.
    \item  The application to those sliding windows to a regression model based on residual deep convolution neural network that uses dilated convolutions, similar to \textit{TCN}\cite{bai2018empirical}.
This architecture was selected as it was successful in many sequence modeling tasks,
outperforming even LSTMs\cite{bai2018empirical}.
\end{enumerate}
The system is trained on a dedicated dataset that contains synchronized sensor and video data for a set of motions typical for a broad domain. For the training the sensor data is filtered and then converted into sliding widows in the same way as the 2D poses and, when needed normalized into  standard range. 
Once trained the system is presented with a video and produces a corresponding sequence of simulated sensor signals within the original standard range. 
%The first step is learning correspondences between 
%sequences of 2D poses and sequences of sensor values. Once this is learned using recordings of generic motions that include both video and sensor values, the trained model can be applied to videos of the target activities to simulate sensor data. Those target videos can be obtained from online repositories or recorded by practitioners and the simulated sensor data generated using them can then be used to train or enhance activity recognition models. In this section we will go over our training pipeline for our method, that is, how we learn the correspondence between video and sensor data. 
%
%\subsection{Training Data Set}

%

\subsection{Obtaining 2D Poses}
\begin{figure}
    \centering
    \includegraphics[width=0.4\columnwidth]{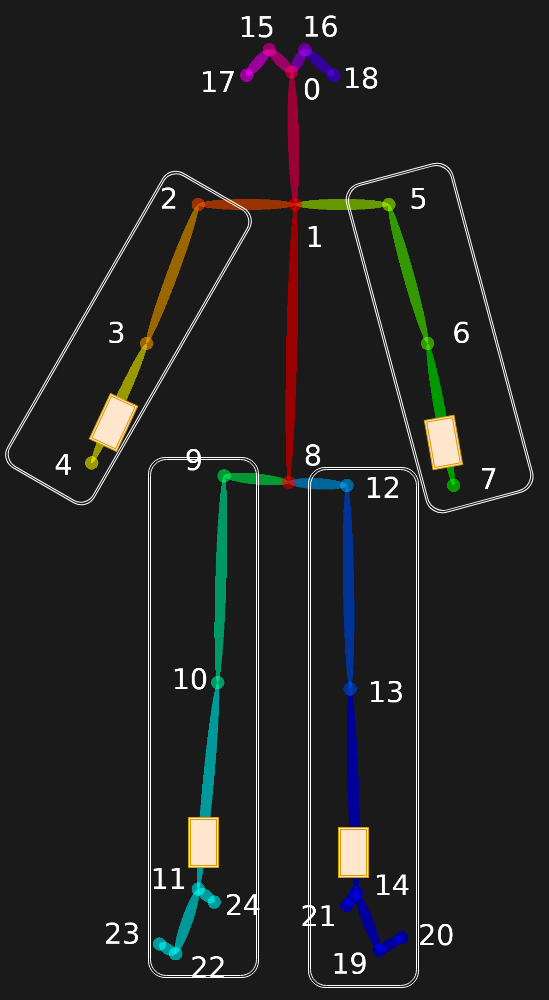}
    \caption{Joints used for each of the $4$ regression models. Those models also receive the scaled hip speed and the change in scale. Orange boxes represent where sensors were placed during the exercises, while their surrounding white boxes describe the joints relevant for each of their regression models. }
    \label{fig:model_separation}
\end{figure}

For obtaining 2D poses we used the same method as in \cite{10.1145/3341162.3345590}, that is, we used the OpenPose library \cite{cao2018openpose,cao2017realtime} to extract frame by frame the 2D poses of subjects. The poses are given in terms of rigid body segments connected by  respective joints a shown in \ref{fig:model_separation}. 
We accept all detected joints from OpenPose with at least 0.0002 confidence. Using those, we track each subject in a video using their Mid Hip coordinates. For each subject, missing joints are filled using linear interpolation. This is done for two reasons: First, not all joints are detected in all frames. Second, videos vary in fps, so we reach $50$ poses per second after linear interpolation. 
This is used for all video sources as the target YouTube videos we used vary in fps from $24$ to $60$, but the training ones we recorded had $50$ fps.
\subsection{Data Preprocessing}
\label{sec:preprocessing}
As indicated above the pre-processing has two aims. First is translating the poses  given in image frame of reference and in pixels as "units", which are specific to a specific image and camera position,  into a more general format. Second is the question to what degree signal processing techniques such as filtering and   normalization can improve the quality of the generate sensor signals.
\subsubsection{Pose Translation and Processing}
The output that we can get from  the pose extractions are, for every video frame, the  pixel coordinates of the individual joints (and/or the angles between the respective rigid body segments). Obviously the  values depend not only on the motion of the user, but also on the camera perspective. The same posture at the top left corner of an image will produce different coordinate values than the same posture at the  the bottom right. For a given motion the position difference between two frames expressed in pixels will be different when executed close to the camera than when executed far from the camera (and will also vary depending on the angle and of course the resolution of the video). On the other hand the sensor values that we want to generate  are  expressed in absolute physical units and are  related only to the user action (and of course sensor placement on the body). 
In summary we need to address  (1)  the translation of the poses into coordinates that are independent of the position of the user in the image and (2) a scaling making the magnitude of pose changes between frames invariant with respect to the size of the user's body in the image.  
Given the power of modern deep learning based computer vision systems one could in principle attempt to have the system learn the corresponding transformation from data. However this would require a very large amount of data on top of the data needed to learn the dependence of the motions observed in the video and the sensor signals as such. As a consequence we have developed the following approach to achieve the required invariance in a pre-processing step before feeding the data into the regression model.     

 %As we want the coordinates to be relative to real distances across recordings, taking into account different camera positions and subject’s heights, we are going to use a scale based on subject joint distance. 
%We will use this value in our preprocessing pipeline
%to take into account part of the variability
%present in those videos, by using it to scale
%our image coordinates.

%In order to better explain the motivation, imagine
%a person of a certain height, unknown to our pipeline. If this person is far away from the camera, the distance in pixels between his joints is smaller than if he was closer. In other words, there is a different scale translating pixel distances to the real life one. On the other hand, the ratios between joint distances should be the same in both cases and thus we can account for this difference by scaling joint positions in each frame by a reliable joint length.

%Another property we want is translation invariability.
%For example, waving your arm on the right of the camera
%should provide the same sensor values as doing the same motion on the left of it. 

To address the first problem we translate the joint position  into a frame of  reference centered in the MidHip joint  the Figure (joint 8 in Figure \ref{fig:model_separation}).  That is, for each joint other than the MidHip compute its new x coordinate at time t ($NewJoint_x^{t}$)
and y coordinate at time t ($NewJoint_y^{t}$) using
    \begin{equation}
    \begin{split}
        NewJoint_x^{t} = scale(Joint_x^{t} - MidHip_x^{t}, scale_t) \\
        NewJoint_y^{t} = scale(Joint_y^{t} - MidHip_y^{t}, scale_t)
    \end{split}
    \end{equation}
Notice that a single scale is used for both x and y, as the aim is to match real space and not to do standard machine learning scaling. In other words, the scale translates pixel distances to joint relative ones,
so it should be the same for both axes in order to
not deform the poses.

As a scaling factor to achieve motion amplitudes independent on the distance from the camera we use body dimensions in the picture. Obviously these are subject to the same camera distance dependent variation as the body parts motions. In other words we express the magnitude of motions in "units" of body  size as seen in the image. To this end  we selected the euclidean distance between Neck and MidHip (joints 8 and 1 in Figure \ref{fig:model_separation}) as a scaling factor for the motions expressed in pixels. Since the distance between the camera and the user can change we  recompute the scaling factor for every frame. Let's call the euclidean distance at time $t$ the $dist_t$. In order to avoid scaling outliers, we are going to use a sliding window of size $3$ seconds, $1.5$ back and $1.5$ forward, and compute the median. This value we call the $scale_t$ which is 
    \begin{equation}
        scale_t = median(dist_{t-75}, \dots ,dist_{t+75})
    \end{equation}
and the function to scale any value $v_t$ is
    \begin{equation}
        scale(v_t, scale_t) =  -1.0 + \frac{v_t} { scale_t} * 2.0
    \end{equation}
 
Using joint positions relative to the body obscures motion of the body as a whole (e.g. moving forward closer to the camera), which also has influence on the sensor signals. 
Fortunately, we can retrieve this information by giving the model access to $2$ extra value related to the $scale_t$ representing the relative speed of scale change \begin{equation}
    speed_t = \frac{scale_{t+1} - scale_{t}}  {scale_{t}}
\end{equation} and the first derivative of this scale speed.  As we explained, changes in scale can be a proxy for absolute movement as the subject is coming closer to / moving away from the camera regardless of specific initial coordinates.

\subsubsection{Additional Signal Processing and Selection}
Signal processing beyond the translation and scaling of the pose values described above can be considered in two categories. The first is further improvement of the quality of the pose information extracted from video.  Here smoothing of the scaling  using the median to remove  small shifts in joint position  and  removing outliers in cases where the network predicts a distorted pose in some of the frames have proven to be most effective. Note that this steps are applied to the video signal only.  With respect to the sensor signal the only prepossessing we have found to be consistently effective  was linear interpolation of  the sensor data to $50$ values per second to match the video frame rate. 

Second is the removal from the signal of components that, for reasons described in Section \ref{sec:problem}, are unlikely to be reliably translated from video to sensor signals.  In essence the idea is to  entirely remove certain types of information from our classification process opting to have less, but more reliable information. Examples are:
\begin{itemize}
    \item Computing the norm of 3D sensor signal.  As explained in Section \ref{sec:problem} rotation of an IMU with respect to the vertical axis determines the distribution of the gravity component between the three axes of an acceleration signal. Similarly when considering the signal in the 
    \item Gravity removal ("linear acceleration"). 
    \item Low pass filtering.  Low pass filtering was motivated by the insight (see Section \ref{sec:problem} point 3) that phenomena such as high frequency "ringing" of the signal caused sudden, strong impact are  fundamentally invisible  in the video data. This means that the simulated sensor data generated from videos will not contain such "ringing".
\end{itemize}
In this above cases the processing needs to be applied to both the sensor signals and the video signals and done during all:  the training of the regression model, the generation of the synthetic training data and the classification of real sensor data with the model generated using the simulated data. Obviously in cases where the information is not present in the video signal at all it is sufficient to remove it from the sensor.

%In this context we have in particular  studied looking on the norm of 3D singals (e.g. norm of the acceleration signal instead of using the x,y and z components individually),   low pass  filtering with various cutoff frequencies and various normalization strategies. 

%The IMU values we are going to use are the accelerometer and gyroscope norm for sensors in the wrists and ankles.

\subsection{Regression models}
\begin{figure}
    \centering
    \includegraphics[width=1.0\columnwidth]{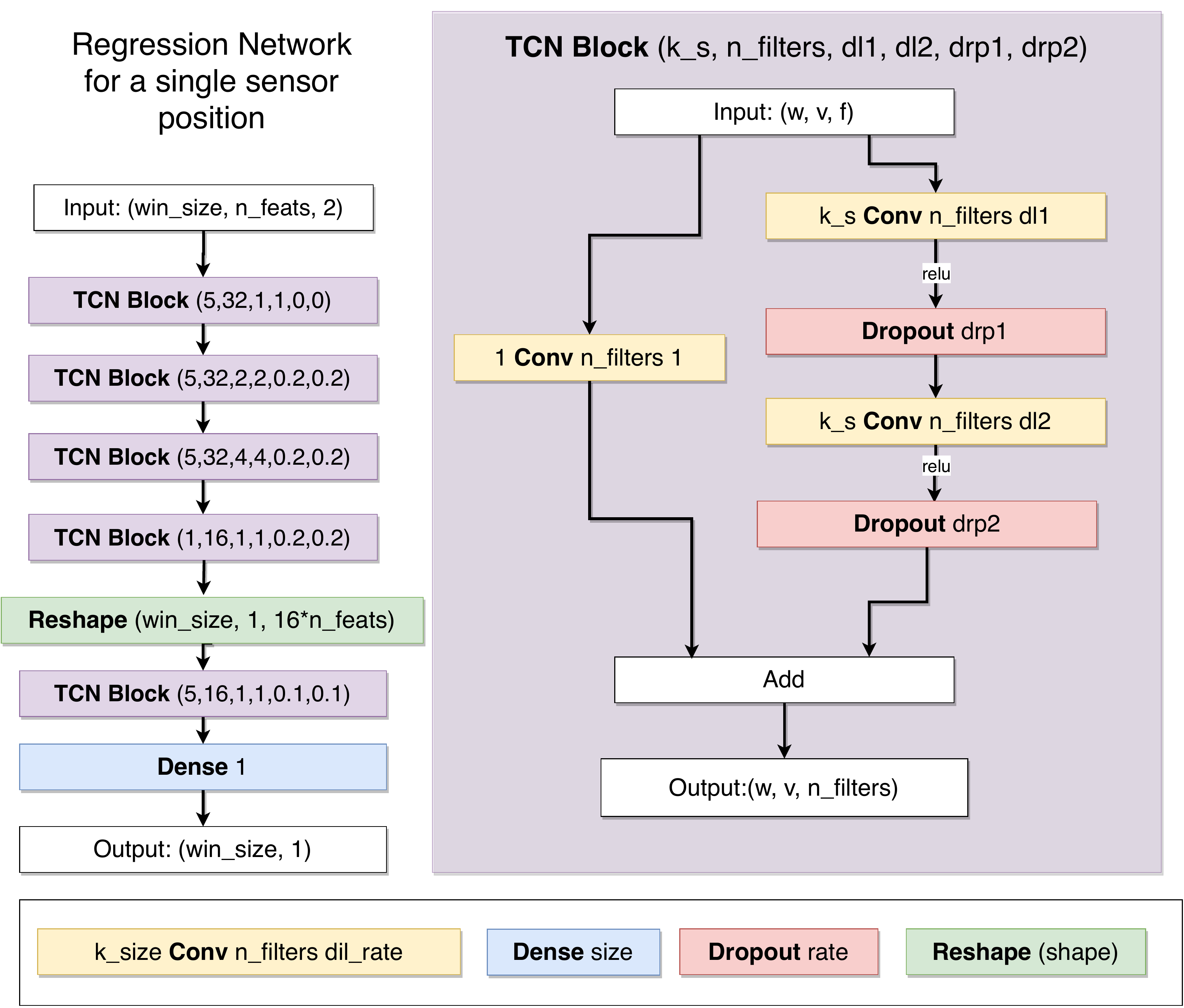}
    \caption{Architecture for the neural network used for regression of a single sensor. }
    \label{fig:regression_model}
\end{figure}

As regression model we selected a residual deep convolutional neural network that
uses dilated convolutions, similar to \textit{TCN}\cite{bai2018empirical}.
This architecture was selected as it was successful in many sequence modeling tasks, outperforming even LSTMs\cite{bai2018empirical}. Our architecture, which can be seen in Figure \ref{fig:regression_model}, relies on dilated 
convolutions and residual connections present in our TCN blocks. Residual connections help train deeper networks by adding the output of convolutions to an identity function\cite{he2016deep}, while dilated convolutions increase the receptive field by orders of magnitude without increasing the computational cost. 
By employing same padding in all convolutions we maintain the temporal size of the inputs, allowing us to go deeper even when using small temporal windows and map the sequence of poses to the sequence of sensors signals. In order to prevent overfitting, we apply dropout in all TCN blocks except the first and reduce the number of parameters by decreasing the number of filters in the 4th TCN block by applying convolutions with filter size 1.
As mentioned before, we train one regression model per sensor position and feature. Each model was trained using the mean squared error loss and the Adam optimizer \cite{KingmaB14} with $0.001$ learning rate and $0.9$ and $0.999$ for
$\beta_1$ and $\beta_2$, respectively. The model was trained for $500$ epochs with early stopping using a patience of $25$ to avoid overfitting. 

{}
It is important to discuss the motivation for our training regimen.
A key question for the training is which information from the video to include in training which sensor. The problem is that:
\begin{enumerate}
    \item Providing information about the motion of body parts that we know, from physical consideration, to be irrelevant for the signal of a given sensor at a given location will "confuse" the model.  Eventually, given enough data, good models will likely learn to distinguish relevant from irrelevant information. However, getting rid of confusing information is known to significantly reduce the amount of required  training data and speed up training. 
    If we had at our disposal the full, exact 3D 6DOF coordinates of the body part (in world coordinates) then, obviously, that would be all the information that would be needed (as the sensor values are fully determined by the 6DOF and their temporal evolution). However, as described in Section \ref{sec:problem}, 2D videos do not contain that information and the missing parts have to inferred from the image semantics, especially the knowledge of bio mechanics. Thus for example consider a person holding the wrist in front of his chest with the camera shooting from the front. Clearly the distance between the chest and the wrist (or the wrist and the camera) is not directly visible. However, looking at the angle of the elbow and the shoulder together with the vertical angle of the chest it can be inferred from approximate human body dimensions.  This means that all joints that determine the position and motion of a body mounted sensors should be included in the input to the regression.    
    \item Providing information about body parts that, from a bio mechanical point of view, do not influence each other on the other hand can be counterproductive and lead to overfitting the specific training motion sequences and combinations. As an example consider the motion of the left and right wrist. With the exception of some very extreme motions (e.g. extremely strong shaking of one wrist making the whole body shake), the motion of one wrist has no influence on each other. However, in many motion sequences there are sometimes strong correlations between the motion of both wrists. The best example is walking when people often swing their arms in sync. When training the system to recognize walking, we want the system to learn such correlations. However, when training a regression that should map motions in an image onto sensor data for arbitrary activities based on physical constraints only this would be undesirable overfitting of artefacts of the specific training sequence. 
    \item Finally there is the question if a joint model should be trained for all sensor signals, if separate models should be trained for all signals from one on body location or if we should train a separate model for each signal at each location. Training one model for all locations and signals  is not advisable due to the same considerations described in the previous point: the need of avoiding overfitting a specific training set due to the system learning spurious correlations between signals from different locations. In principle, training a model to generate all the signals from a given body location should allow it to capture the physical dependencies between those signals and should thus improve the quality of the results. However, in our experiments we have found  individual models trained for each sensor modality at each location to perform best. This may be due to insufficient training data. It may also be necessary to introduce onto the model mechanisms for explicitly encoding physical boundary conditions as propose e.g by the concept of Physically Informed Neural Networks \cite{raissi2019physics}.
\end{enumerate}
 Thus for each target sensor signal we  trained one regression model per possible sensor placement using only the joints relevant for it. We placed sensors on the wrists and calves, as shown in Figure \ref{fig:model_separation}, which also shows which joints are used for each position. For each one of those inputs we built a regression model as depicted in Figure \ref{fig:regression_model}. Regression is done using a window of size $16$ frames (around a third of a second) and step $1$. This increases training data and was also selected to avoid overfitting specific movements. The regression output is $16$ numbers representing the IMU values for one channel at those times. The input for each regression model using $n_j$ joints is a tensor $(16, n_j + 2, 2)$, that is, the 2D coordinates of the joints involved and also the $2$ extra values related to the scale mentioned earlier.
 For the joints involved for each position, see 
Figure \ref{fig:model_separation}.
\subsection{Datasets for Training the Regression}
As already described our aim is to train a model that can generate simulated sensor data not just for a single specific set of activities but for a broader domain. Thus having our regression model users should be able to pick their own activity sets that they want to recognize and use our model to generate the required training data from online videos that are easy to acquire and label.

The idea is to train the  regression model on a set of motions that are representative "basic components" of the activities of the respective broader domain.  The core question is how to define the "broader domain" and how to identify the corresponding "basic components".  For  this work we have selected the broad area of aerobics like physical training as our domain.  It is a very broad  domain with a great variety of different exercises for which online videos are extremely popular. In terms of applications the ability to monitor the exercises is something that current fitness trackers lack despite the popularity of such exercises and the associated potential large user base.
In technical terms the domain has the advantage that the corresponding videos tend to provide a stable, easy to process frontal perspective with the relevant parts being clearly visible most of the time (as the whole purpose of the videos is to show the viewers how to move). The activities tend to be characterized   by distinct motion of the limbs and posture changes which means that recognition systems should be reasonably robust with respect to senor data noise and inaccuracy. 

\begin{figure}
    \centering
    \includegraphics[width=1.0\columnwidth]{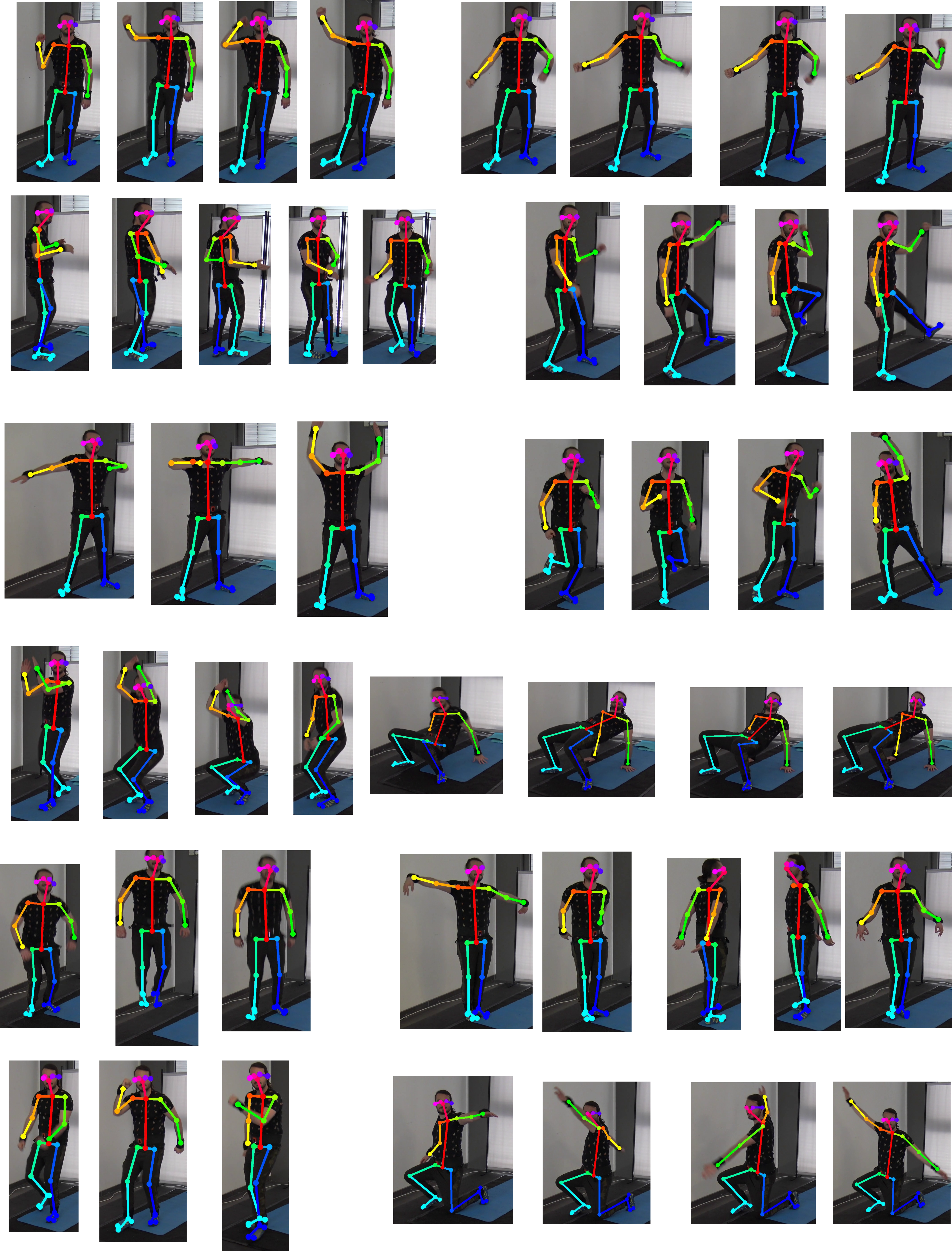}
    \caption{Example of a subject performing some of the movements present in our generic motions dataset used for training the regression model. }
    \label{fig:ex_generic}
\end{figure}

With respect to the "basic components" of the motions we have first started by showing the  participants examples videos of the domain and asking them to perform   random motions  related to those videos. This approach has not produced good results ,however, as people tended to repeat the same not necessarily representative motions. We have then turned to high school video  material explaining the degrees of freedom of human joints and the types of motions that different muscles can cause.  We cut together  parts of $7$ of those videos into an $11$ minute compilation\footnote{Dear reviewer, you can see this video in the extra materials. Public YouTube link will be added here if the paper is accepted.} and had $8$ subjects try to repeat the motions from the videos while wearing sensors on the locations for which we wanted to train our regression model (wrists and lower legs). We filmed each user in a different day and our camera placement and angle changed across recordings, but was always facing the corner depicted in Figure \ref{fig:ex_generic}, where we can see examples the generic motions.
Regarding sensor synchronization, we use the same approach as in \cite{10.1145/3341162.3345590}. After starting the camera recording, we turn on all sensors and perform our synchronization gesture, which consists of holding all IMUs stacked on top of each other and moving them together up and down in front of the camera. This is done in order to later bring all sensors to video time using the acceleration peaks and their corresponding video frames.
After the synchronization gesture is performed at least $3$ times, we start recording the
generic motions. During a user session, subjects were instructed to try to follow in real time the motions present in our compilation video, which is playing in a monitor nearby. In order to increase the variability of motions, we told them that it was more important to move than to perform the motions correctly. For example, when quick motions happen in the videos, it is better to move quickly even if one cannot follow the specific quick motion. Only one of the subjects performed those seed motions more than once. His second session was selected for validation, while all others were selected for training the regression. 

\section{Evaluation}\label{sec_collection_evaluation}
With respect to evaluation it must first be noted that our aim is not to generate signals that are as close as possible to the signals that real sensors would generate when performing the respective motions. Instead  we want to facilitate the generation of training data that will allow activity recognition systems to achieve performance that is as close as possible to the performance achievable with training data recorded with real sensors. Obviously these goals are related, but not identical.  Generating sensor data that is very similar or even identical is certainly a sufficient condition for getting recognition results that are close to what is achievable with real sensor based training data.  However it is not a necessary condition. Thus, if the system can replicate enough distinct features of the signal to separate the classes in a particular application then the fact that it may miss or fail to reproduce other signal features is of no significance.  This is illustrated in Figure \ref{fig:moresignals}. The signals shown in the Figure are the actual signals we used.

As a consequence we focus on the evaluation of the performance of activity recognition trained using simulated sensor data generated by our system from respective videos and provide limited analysis of the quality of the generated sensor signal as such (see Section \ref{sec:signalquality}). The remainder of this section describes the dataset and approach used for the evaluation. The results are presented and discussed in the next section.

\begin{figure}
    \centering
    \includegraphics[width=1.0\columnwidth]{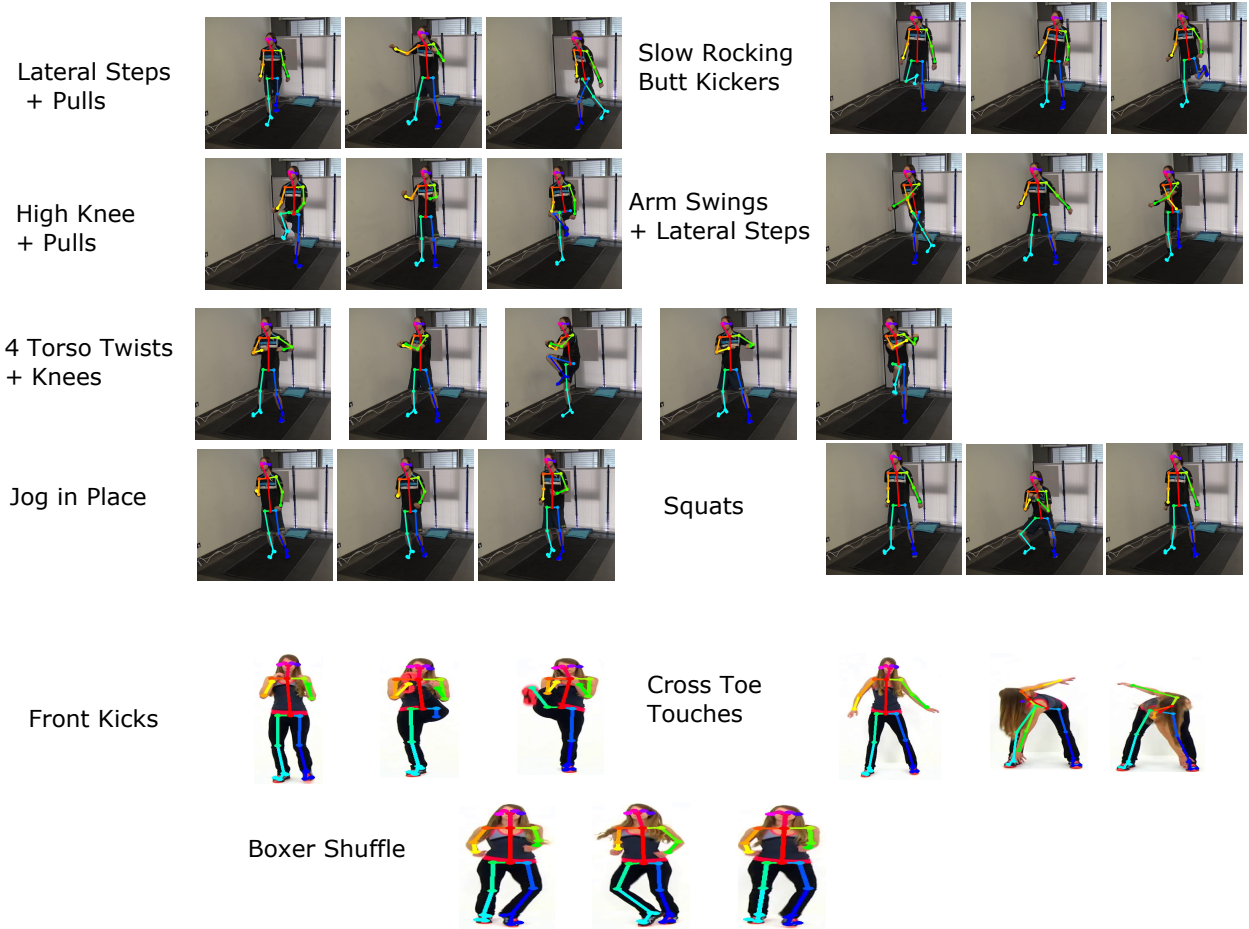}
    \caption{Short depiction of the motions present in the $10$ exercises of the Drill dataset, in which participants performed the exercises of \url{https://www.youtube.com/watch?v=R0mMyV5OtcM}.
    The first examples are of one of our subjects, while the last are from the original YouTube video each subject tries to replicate. In this dataset, each exercise is repeated once for $30$ seconds.}
    \label{fig:ex_ex}
\end{figure}

\subsection{Data Collection for Activity Recognition Based Evaluation}
As already mentioned in Section \ref{sec:problem} we have selected the broad domain of aerobic like physical exercises for our evaluation since it is an interesting and relevant application area while at the same being well suited for our approach. 
We selected a  set of activities $10$ we refer to as the  "Drill dataset" and that we have already used in our preliminary work described in  \cite{10.1145/3341162.3345590}. Those are part of a light cardio workout taken from a popular YouTube fitness channel \footnote{ \url{https://www.youtube.com/watch?v=R0mMyV5OtcM}}. The specific exercises can be seen in Figure \ref{fig:ex_ex}. Altogether we have collected  $12$ YouTube for videos of people doing those exercises including the one used in \cite{10.1145/3341162.3345590}, which is the only one that contains all target exercises. The full list of videos used can be seen in Table \ref{tbl:yt}. In the videos individual exercises are performed sequentially so that  to label a video, one only needs to provide for each one the frame intervals for each activity present.  Overall the videos provide us with about $24$ minutes of usable footage of the 10 activities.  
In addition to the online video material we have recorded data with volunteers at our lab. To this end we have equipped them with IMUs (XSENSE) fixed to their lower legs and wrists (see Figure \ref{fig:model_separation} for the sensor placement)
and asked them to imitate the exercises in the video. We have recorded the subjects on video with a camera perspective similar to the online videos to ensure that we have not only real sensor data but can also generate simulated data for the subjects using our regression model.
Overall we had $28$ subjects performing the $10$ exercises.   
Of those  we have $17$ users who performed a single session only, which are considered the training set.  The test set consists of the $11$ users who performed more than one session ($5$ performing $2$ sessions, $4$ doing $3$ and $2$ performing $4$ for a total of 30 sessions). This gives us a diverse training set, a large number of testing sessions and the ability to test on the most difficult scenario: the training and testing sets being different users. For an overview of the datasets used and the length and number of users in each one, see Table \ref{tbl:srcs}.

%For each one of those intervals, the whole preprocessing step is done and then windows of size $16$ and step $1$ are generated. In total, those segments containing the target classes reach $14.28$ minutes.  If we sum the sizes of all windows we generate from them we have roughly $24$ minutes of useful time.

\begin{table}[]
\caption{YouTube videos used to generate artificial sensor data. Each video has a single subject performing one or more activities and thus we have $14.28$ minutes of the target activities in total. }
\begin{tabular}{llll}
\hline
Video URL                                   & Ativitie(s)               & Seconds Used & fps \\ \hline
https://www.youtube.com/watch?v=7X2Yx29DdBY & Cross Toe Touches         & 113          & 30  \\ 
https://www.youtube.com/watch?v=8gLdmb9Ivkw & Lateral Steps + Pulls     & 32           & 30  \\ 
https://www.youtube.com/watch?v=9-jBOcGeQcg & 4 Torso Twists + Knees    & 16           & 30  \\ 
https://www.youtube.com/watch?v=afghBre8NlI & Squats                    & 83           & 30  \\ 
https://www.youtube.com/watch?v=enz5TSRMmyM & High Knee + Pulls         & 13           & 25  \\ 
https://www.youtube.com/watch?v=g-S1c-Scu3E & Lateral Steps + Pulls     & 33           & 30  \\ 
https://www.youtube.com/watch?v=Kn621fAVEEI & Boxer Shuffle             & 9            & 30  \\ 
https://www.youtube.com/watch?v=MG8DJpN-35g & 4 Torso Twists + Knees    & 48           & 30  \\ 
https://www.youtube.com/watch?v=mGvzVjuY8SY & Squats                    & 100          & 30  \\ 
https://www.youtube.com/watch?v=oMW59TKZvaI & Slow Rocking Butt Kickers & 10           & 24  \\ 
https://www.youtube.com/watch?v=ZiJdpPJbqYg & Front Kicks               & 4            & 30  \\ 
https://www.youtube.com/watch?v=R0mMyV5OtcM & All of the Drill Activities              & 303.81       & 60  \\ 
\hline
\end{tabular}
\label{tbl:yt}
\end{table}
\begin{table}[]
\caption{Details about the datasets used. Every drill session lasts $5$ minutes, while every the generic motions last $11$.
For our generic motions, we use one session of each subject for training the regression and the second session of one for validation. For the Drill dataset, we use users with a single session as the training classification set and those with more than one as the test set. No user is shared between any of the datasets. }
\begin{tabular}{cccccc}
\hline
Dataset                & video & sensor data & Subjects & fps      & Total minutes \\ \hline
Collected generic motions & yes   & yes         & 8        & 50       & 99            \\ 
YouTube videos         & yes   & no          & 12       & 24 to 60 & 14            \\ 
Drill Dataset Training & yes   & yes         & 17       & 50       & 85            \\ 
Drill Dataset Test     & no    & yes         & 11       & -        & 140           \\ 
\hline
\end{tabular}
\label{tbl:srcs}
\end{table}

\subsection{Evaluation Procedure}
\label{sec_eval_proc}
%The evaluation procedure is similar to the one performed in  \cite{10.1145/3341162.3345590} and is as follows: We first order the training users from best to worst as determined in \cite{10.1145/3341162.3345590}, that is,
%by their mean F1 score in the test set using a deep convolutional neural network.
The evaluation procedure compares various ways of training the system with simulated sensor data generated using our regression model with the baseline of a system trained using real sensor data.  For the baseline we use, as described above, the sensor data from the volunteers in our Drill data with the $17$ users that have recorded one session only being used for training and the remaining $11$ users with their total of 30 sessions as testing set. For the generation of training from videos  using our regression model we have the $12$ YouTube Videos and the videos of the volunteers of our Drill experiments. This way we test our regression model on activities which it has not seen before (as it was trained on a collection of more or less random motions from the high school educational videos as described in the previous section). In summary we consider 3 different sources of training data: 
\begin{enumerate}
    \item The real IMU data from our training users in the Drill dataset. This is the "gold standard". Systems trained with this data should perform best.  
    \item Simulated  IMU  data generated from  the videos of our training users in the Drill dataset. Here the system trained on data generated from video is trained on exactly the same users and activity instances as the baseline system ensuring the "fairest" and most consistent  comparison.   
    \item IMU data generated from the 12 YouTube videos. This is a setup that is likely to be most relevant for real world applications.  
\end{enumerate}
When investigating the value of sensor data generated from videos using our regression model we consider four more specific questions:
\begin{enumerate}
    \item  What is the effect of the various possible processing techniques  described in Section \ref{sec:preprocessing} and the performance difference between a system trained on the real sensor data and one trained on data generated from video using our approach? In this context it is important to also consider the effect of the respective techniques on the absolute performance of the baseline system.  Thus, we do not want to consider techniques that may reduce the difference between simulated and real sensor data based systems at the price of making both systems significantly worse.   
    \item  How well does our approach perform for different types of sensor signals? The question of sensor choice has already been discussed in Section \ref{sec:problem}. Again we need to consider not only how the different sensor choices effect the difference between simulated and real sensor data but also how they impact the absolute performance. 
    \item  Can we compensate for potentially inferior quality of the training data generated from videos by providing a larger amount of training data? As described in the introduction the problem of sensor based HAR is not as much getting training data at all, but getting such data in large amounts. At the same time it is well known that the performance of most ML system, in particular deep neural networks, tends to depend on the availability of large amounts of data.  Thus given that our approach gives the user access to huge amount of data contained in online videos (much more then can realistically be collected as labeled data with on body sensors), it is not necessary to match the performance of real sensor data in tests on  the same sized training sets (as done with respect to points 1 and 2 above). Instead we need to investigate what happens when we provide the system using simulated sensor data more training examples then the real sensor data based system.  
    \item Can the quality of training based on simulated sensor data generated from videos be improved by combining it with a small amount of real sensor data? As already described recording small amounts of labeled sensor data is possible in many applications. The question is if such small amounts of real sensor data can "fine tune" HAR systems trained on large amounts of simulated data to improve their performance.  
\end{enumerate}
In order to assess the quality of our simulated data, we will always test models on the real data of our test users. 
From our previous experiments\cite{10.1145/3341162.3345590} we ordered train users by how good their data alone can be used to classify the test set. In our experiments here, we will vary the number of training subjects,
 selecting them in order, so if we select two training users we are selecting the two best ones, as we know from previous work\cite{10.1145/3341162.3345590} that the quality of real user data varies a lot in this dataset.

%\begin{itemize}
 %   \item A model trained of the real IMU data for those $n$ users. This is our baseline for what can be achieved with that amount of real data.
    %\item A model trained on the simulated data generated from the videos of those same $n$ users. This represents how %well our regression model, that has never seen those specific motions or users, can recover the performance of the %original real data.
   % \item A model trained on only IMU data simulated from YouTube videos, representing how well 
  %  \item A model trained on the simulated from YouTube as well as the real data for those $n$ users, to see how opport
 %   \item real data from $1$ or $2$ of the $n$ users together with the simulated data from the $n$ user's videos.
%\end{itemize}
\subsection{Recognition Approach}
\begin{figure}
    \centering
    \includegraphics[width=0.3\columnwidth]{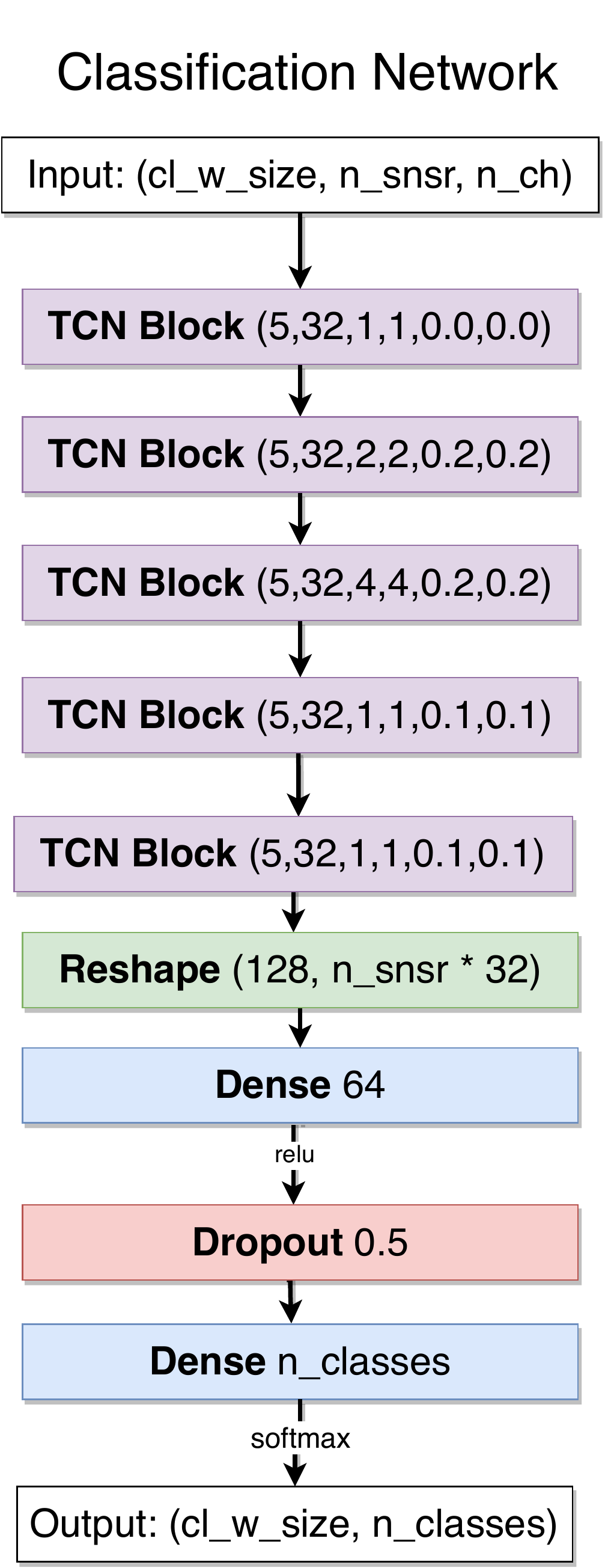}
    \caption{Architecture for the neural network used for classification. For a definition of the blocks see Figure \ref{fig:regression_model}. The size of the window for classification was $128$, representing $2.56$ seconds and the number of classes in our target dataset is $10$. }
    \label{fig:classsification_model}
\end{figure}

Since the  aim of this work is not to optimize the recognition of a specific set of activities but to evaluate the usefulness of simulated sensor data generated from video for wearable HAR systems in general, we use a fairly standard architecture that has proven to be well suited for a variety of activity recognition tasks. The full network architecture can be seen in Figure \ref{fig:classsification_model}, and, just like the regression model, is based on \cite{bai2018empirical}. We also know from previous work\cite{10.1145/3341162.3345590} that this specific network architecture is suited to this specific task.

The input to the recognition system are $2.56$ seconds (128 frames) long windows, each  classified as belonging to one of the $10$ exercises. This window size is sufficient to capture the motions of each exercise, which is repeated many times in a span of $30$ seconds. Of course, one window has data from all available sensors for that test. For example, if we are considering accelerometer norm from all placements then each window is a tensor (128,4,1). If, on the other hand, we are using both accelerometer and gyro norm for all placements, then it is (128,4,2).

Regarding the outputs, notice that all convolutions in this architecture use \textit{same} padding, meaning that it outputs class predictions for each window step instead of one label per window. This allows the network to predict with a better granularity even in cases where two activities are happening inside the same window. When evaluating our system, we will predict window labels by majority voting both for the ground truth and the predicted labels. This is done by sliding the window half the window size (64 frames representing $1.28$ seconds).

We trained the network using the categorical cross-entropy loss function and the Adam optimizer \cite{KingmaB14} with $0.001$ learning rate and $0.9$ and $0.999$ for $\beta_1$ and $\beta_2$, respectively. 
Each model is trained for $500$ epochs with early stopping using a patience of $25$ to avoid overfitting. If any real data is included in a test, a stratified $10\%$ random selection of it is used as the validation set. If no real data is included, the validation set consists of the same selection strategy but applied to all the simulated data. This procedure was done not only to guarantee the quality of the validation set, but also not to give an unfair advantage to configurations that combine simulated and real data. If we simply selected a stratified $10\%$ of all data for validation, tests that combine data simulated and real could benefit from having more real points in the training set.

{}

\section{Results}\label{sec_results_discussion}
\subsection{Signal Level Evaluation}\label{sec:signalquality}
\begin{figure}
    \centering
    \includegraphics[width=0.90\columnwidth]{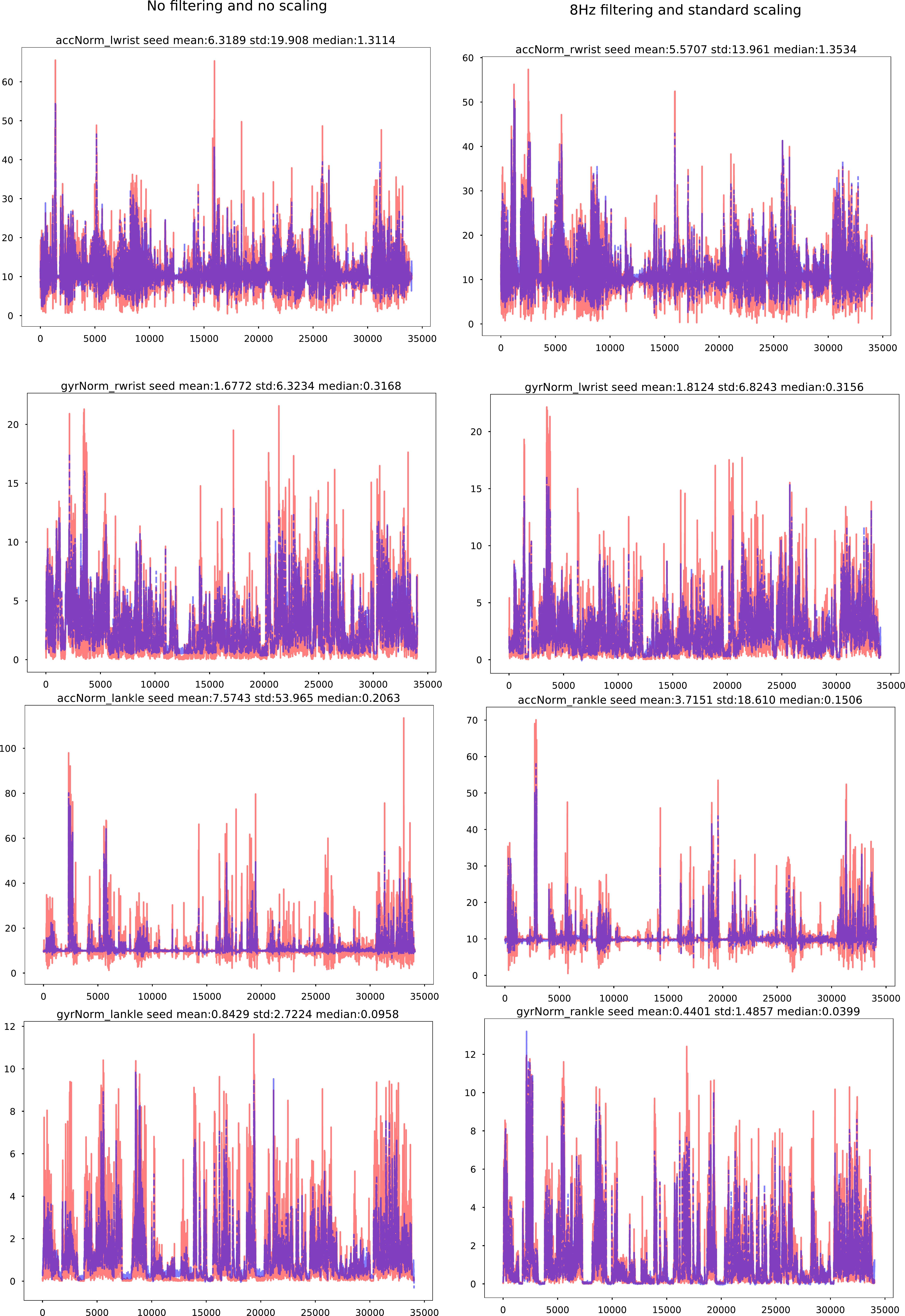}
    \caption{Real signals (red) versus simulated (blue) in the regression training set (generic motions) for the norm of signals}
    \label{fig:moresignals}
\end{figure}
As already explained the aim of our work is not to generate signals that mimic a real sensor signal as exactly as possible, but to facilitate training of activity recognition systems that will later be applied to real signals. In broad terms this means that the key performance indicator is not some sort of numerical  dissimilarity measure but the ability to replicate relevant, characteristic signal features that can be used for class discrimination.  
As the later is difficult to quantify on signal level, we will, as already elaborated,  do quantitative evaluation on the classification tasks with the results being presented in Sections \ref{sec:eval_perf_all} and \ref{sec:class_level_filter}.  Nonetheless it is informative to have a qualitative look at the signals that our system generates for various sensors and how they compare to the real sensor signals generated in the same situation. To this end we consider situations where we have video and senor data for the same activity. We then plot over each other the actual sensor signal (red in all figures) and the corresponding simulated sensor signal generated by our system (blue in all figures).
%Initial examples of the signals generated by our system for the Drill dataset  were already provided in in Figure \ref{fig:regr_examples}. 
We begin by looking at signals generated for the dataset that we used for training the regression. This is in a way the easiest task as we do not have to deal with the question of how representative the training dataset that we assembled for our regression model  is.  In Figure  \ref{fig:moresignals} we first consider examples of signals for the accelerometer and gyroscope norm $\sqrt{x^2+y^2+z^2}$. As explained in Section \ref{sec:problem} this avoids problems associated with the difficulty of estimating small rotations around the limb axis which have influence in particular on the projection of the gravity vector onto the individual sensor axis. The signals are shown for a period of time of around 10min. For each signal we show the unprocessed signal and  a version filtered with 8Hz (the impact of the filtering will be discussed later). General observations from Figure \ref{fig:moresignals} are:
 \begin{enumerate}
 \item Overall the simulated signal has the same trends and large scale features as the original signal. 
 \item Many of the smaller features such as distinct peaks are also matched by the simulated signal. However some are missed or have a much smaller amplitude in the simulated signal. 
 \item The main difference between the simulated and the real signal is in the amplitude. In most cases the simulated amplitude is smaller, however not by a constant factor that could be overcome with simple scaling. The underestimation is particularly pronounced for high frequency peaks. This can be explained by a number of factors. First of all high frequency components ("details") tend to be in general more difficult to reproduce and given the limited size of of our training set some limitations in this area are not surprising. Second, in particular for the acceleration sensors, some of the peaks are due to physical effects which are not present or very difficult to capture in the video (e.g. high frequency "ringing" after sudden impact, see Section \ref{sec:problem}). 
 \item Especially in the acceleration signals there is a "baseline shift" like effect (especially visible in the third row of Figure \ref{fig:moresignals}.  At times the system seems to completely ignore parts that are below a baseline situated just below 10 (corresponding to around 1g).  This can be attributed to downward motions where the earth gravity is subtracted from the acceleration. Thus a free falling (accelerating downwards with 1g) experiences no acceleration force (is weightless). This means that the acceleration norm is smaller then 1g (actually 0 in free fall), at least as 
 experienced by the real sensor. For all other motion on the other hand, the value of the acceleration is always equal to or above , at least in the norm, acceleration. Given the limited size of the training set for the regression model and the fact that we did not include any semantic analysis to detect the "down" direction it is not surprising  that our model struggles to capture this phenomenon. One way of dealing with the problem is to use linear acceleration which can be derived by IMUs, which is discussed below.     
\end{enumerate}
%As we can see in \ref{fig:fft_drill}.
\begin{figure}
    \centering
    \includegraphics[width=0.49\columnwidth]{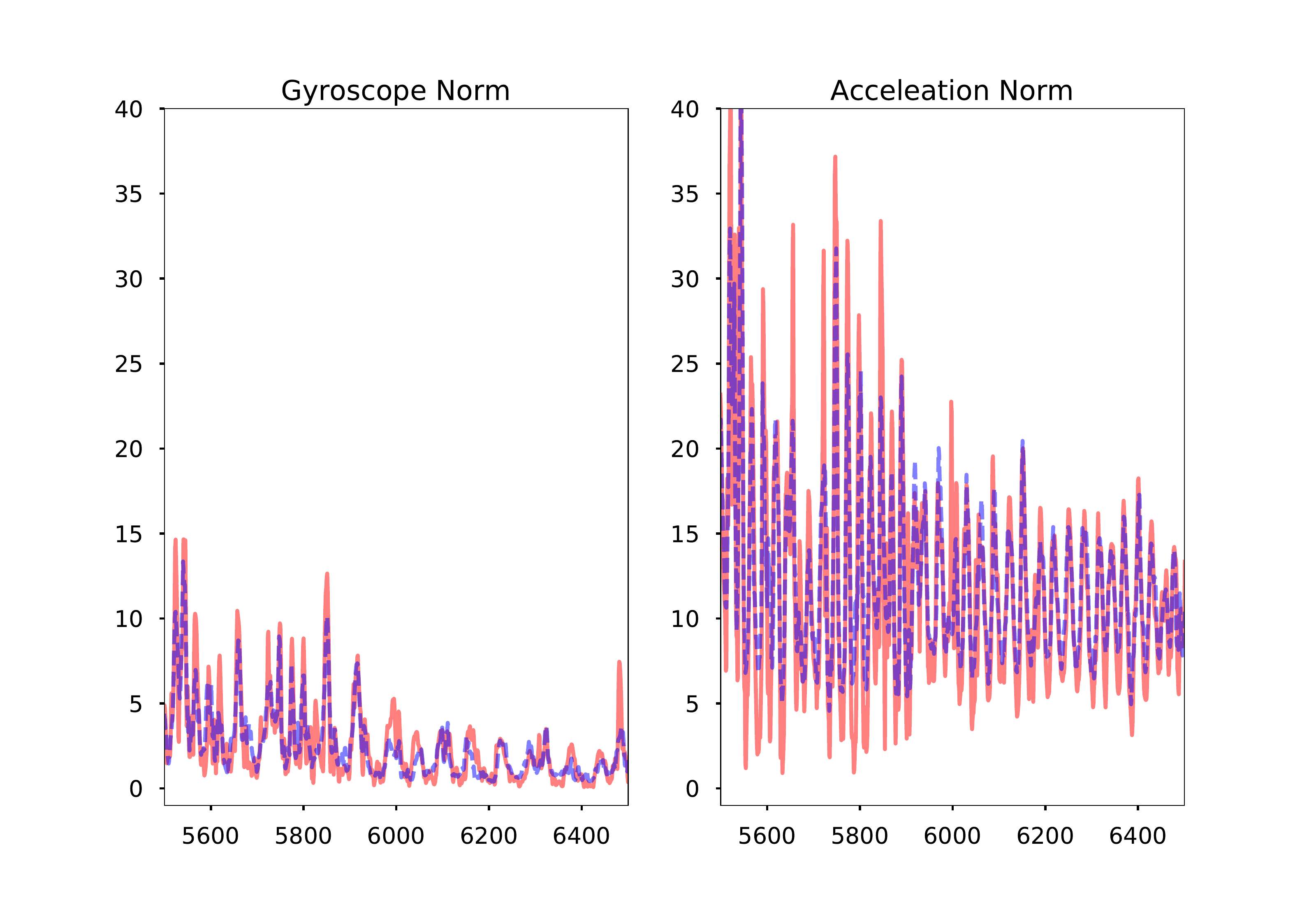}
    \includegraphics[width=0.49\columnwidth]{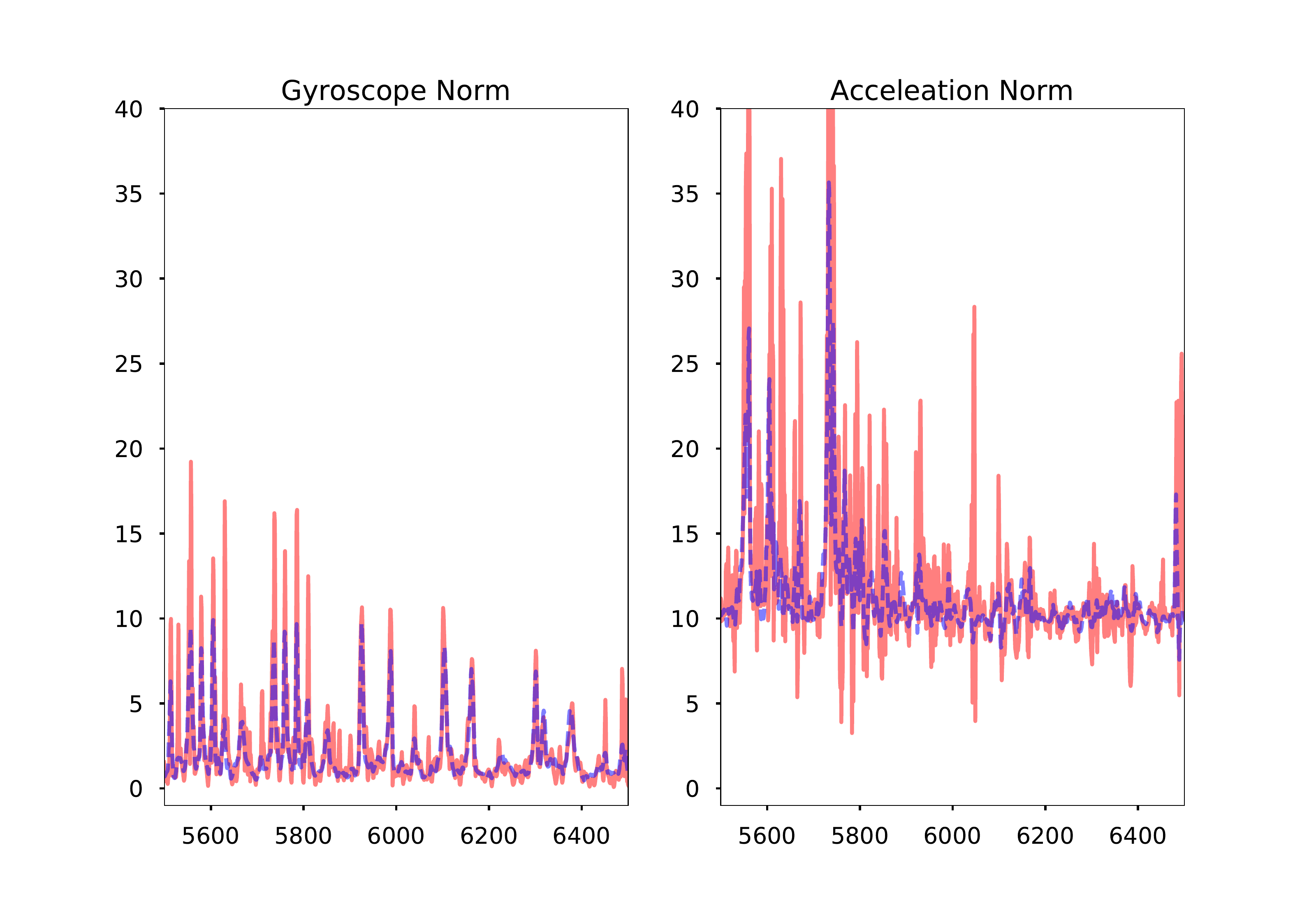}
    \caption{Real signals (red) versus simulated (blue) in the regression training set (generic motions) for acceleration and gyro norm. All cases trained without neither filtering nor standard scaling.}
    \label{fig:acc_and_gyr_signals}
\end{figure}

\begin{figure}
    \centering
    \includegraphics[width=0.49\columnwidth]{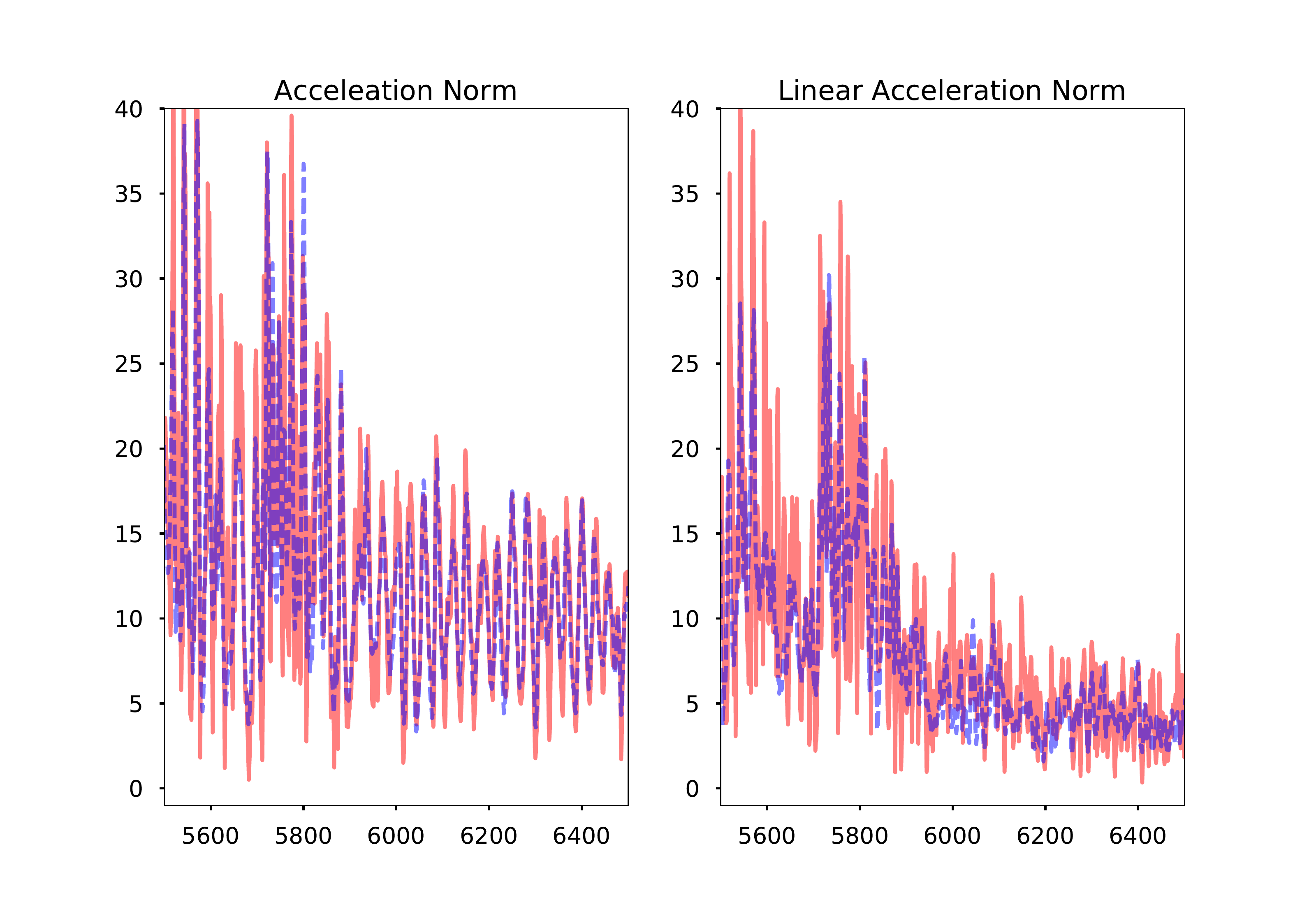}
    \includegraphics[width=0.49\columnwidth]{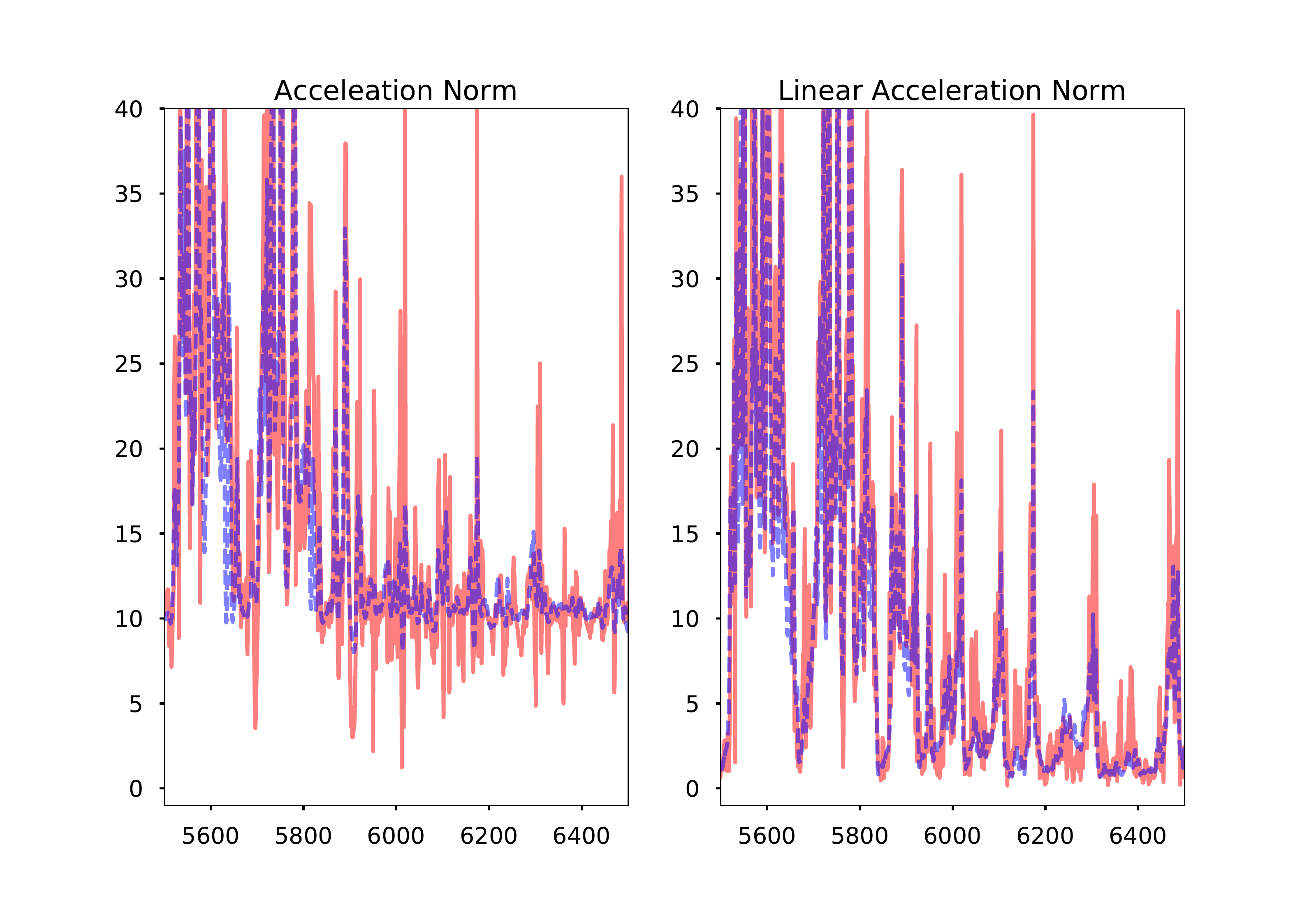}
    \caption{Real signals (red) versus simulated (blue) in the regression training set (generic motions) for acceleration norm, linear and not. All cases trained without neither filtering nor standard scaling.}
    \label{fig:ax_norm_signals}
\end{figure}

To get a better understanding how the system handles detailed signal features Figures \ref{fig:acc_and_gyr_signals} and \ref{fig:ax_norm_signals} show zoomed views of 2 signal segments that were selected to cover the "good" as well as the "bad" cases. Figure \ref{fig:acc_and_gyr_signals} shows for each signal a comparison of the acceleration and gyro norms. To explore in more detail the issue raised in point 4 above Figure \ref{fig:ax_norm_signals}  shows not just the acceleration norm, but also the norm of the linear acceleration which excludes the gravity contribution. The main observation are:
\begin{enumerate}
    \item The left part of Figure \ref{fig:acc_and_gyr_signals} shows an example of a "good" simulation.  It can be seen that the simulated gyro signal nearly perfectly tracks the real gyro, even through fairly subtle features.  Except for the amplitude issues the same holds for the accelerometer signal. Note that while the signal segment contains subtle and fast components, it does not have very high frequency "singular" peaks, which is a key reason why the system works so well here. 
    \item On the right side of Figure \ref{fig:acc_and_gyr_signals} a case of much poorer performance is shown. This is largely due to the presence of many singular high frequency, high amplitude peaks which especially the acceleration signal fails to track correctly. 
    \item With respect to the linear acceleration Figure \ref{fig:acc_and_gyr_signals} shows two things. First we see on the right side that using the norm linear acceleration instead of the norm raw acceleration signal indeed does solve the problem of the system refusing to model signal values below the baseline of 1g.  On the other hand, as shown in the left part of the figure, using linear acceleration may lead to signal characteristics changing and acquiring additional  high frequency peaks which (as already discussed) can be difficult to model.  
\end{enumerate}

\begin{figure}
    \centering
    \includegraphics[width=0.49\columnwidth]{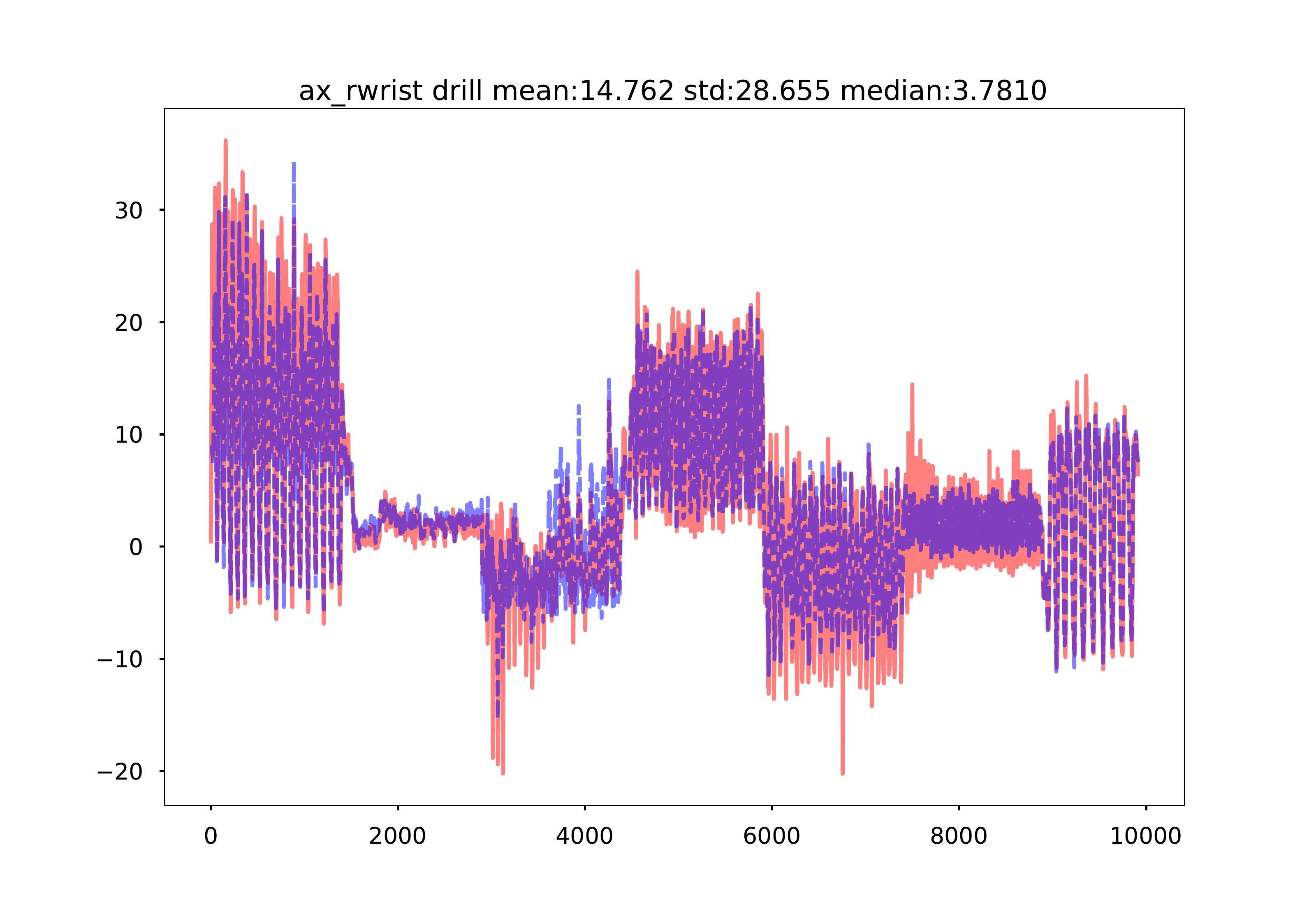}
    \centering
    \includegraphics[width=0.49\columnwidth]{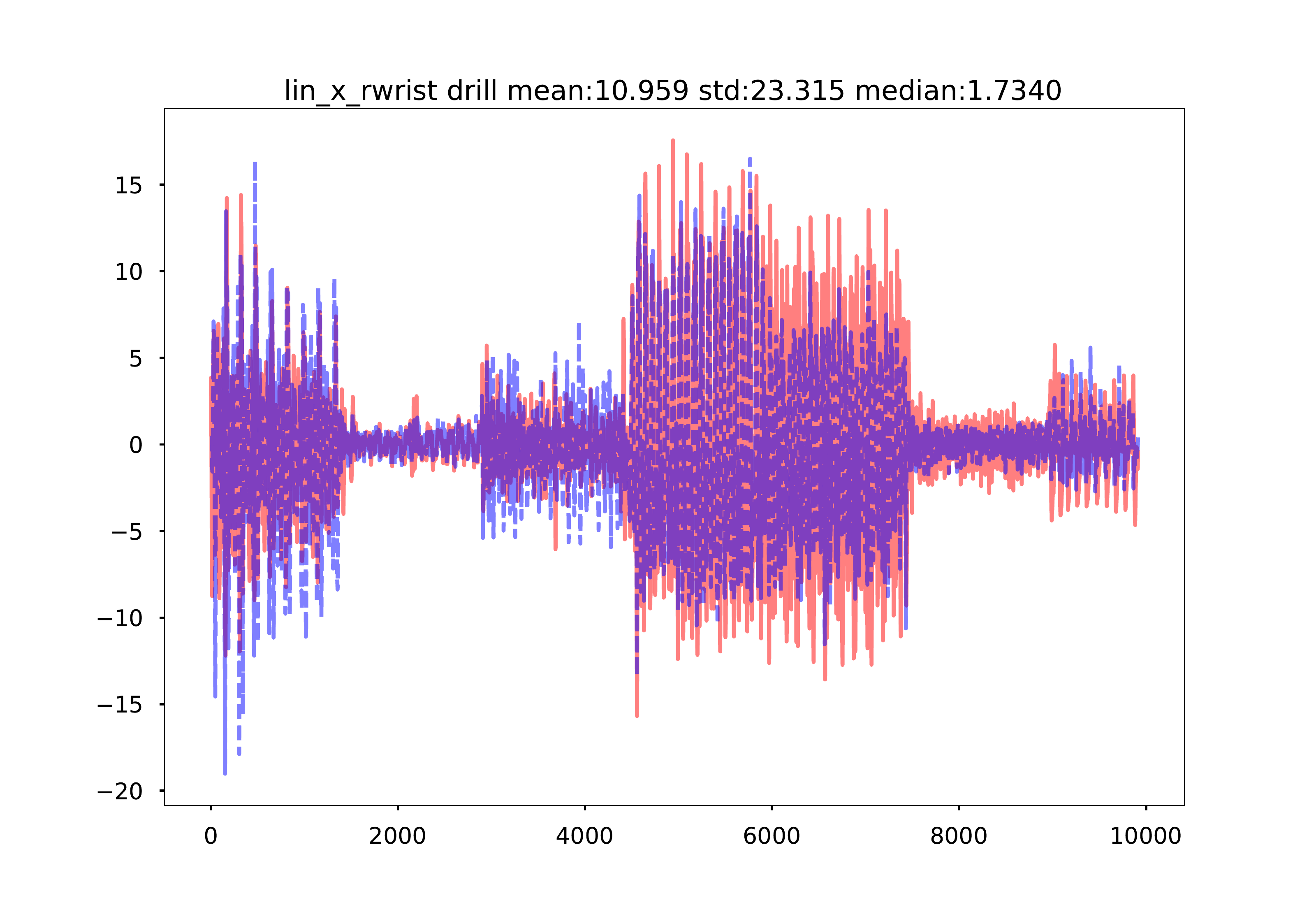}
    \centering
    \includegraphics[width=0.49\columnwidth]{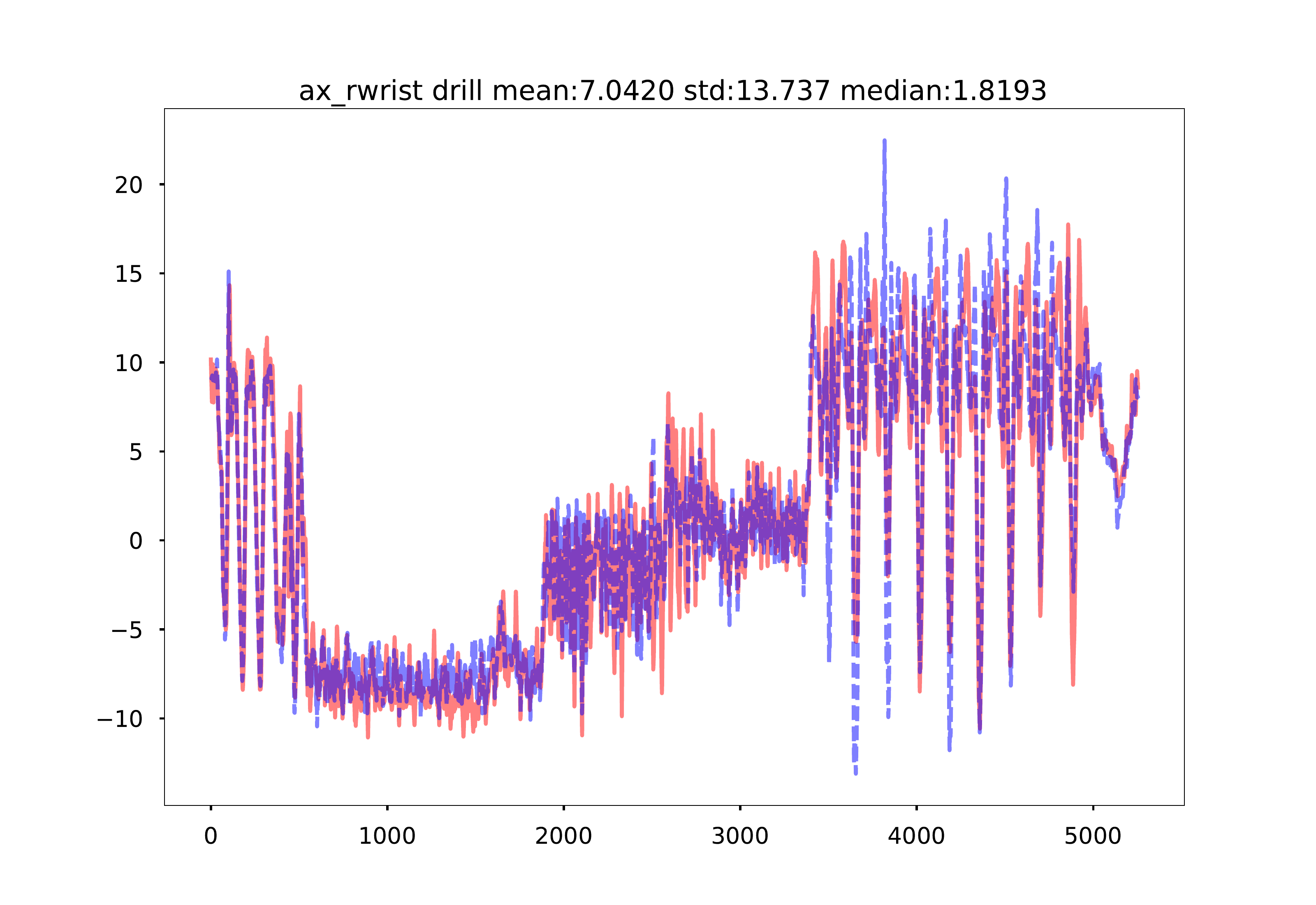}
    \centering
    \includegraphics[width=0.49\columnwidth]{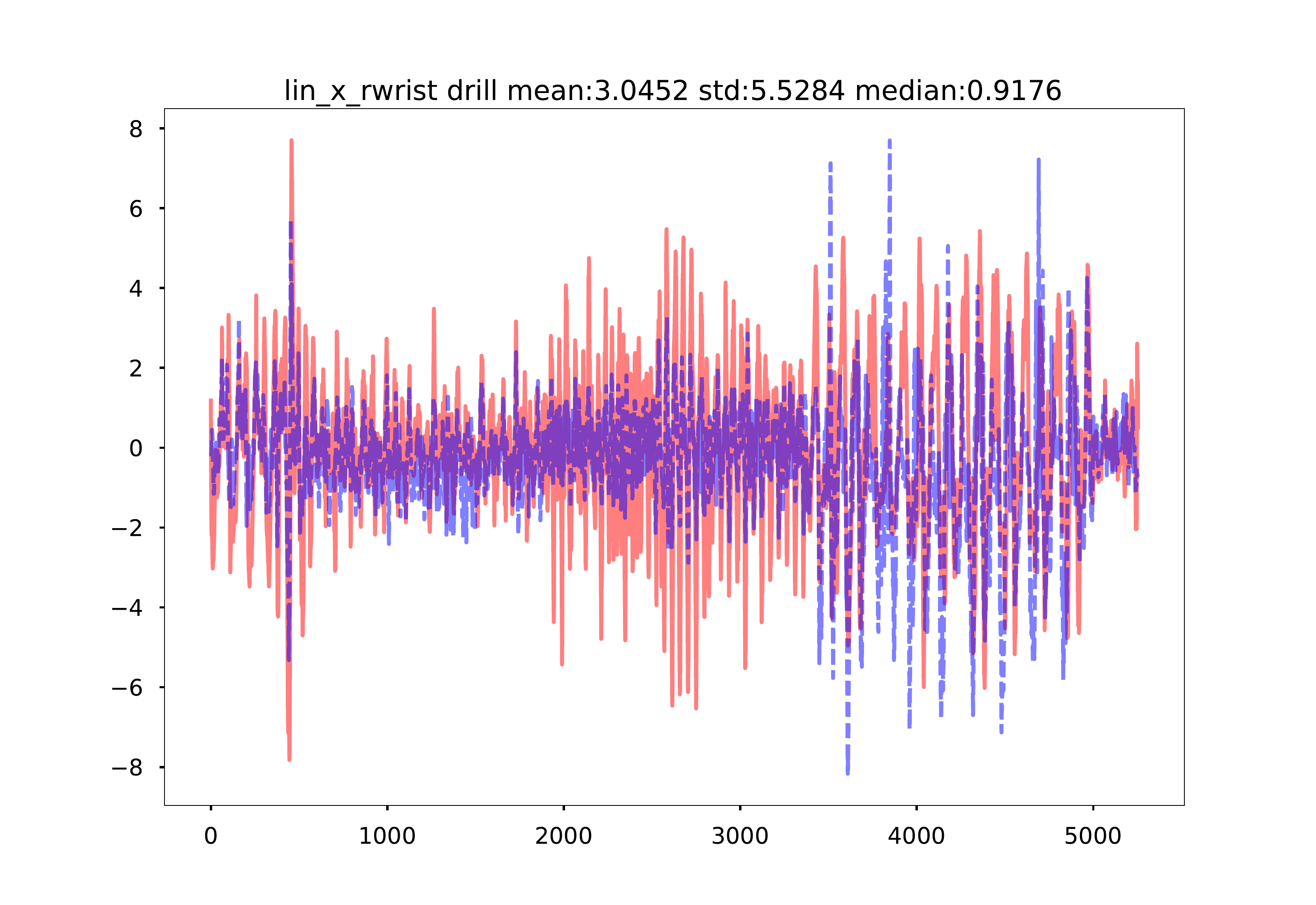}
    \caption{Examples of the signals generated by our system in the target dataset (fitness exercises). Specifically we show the acceleration parallel to the lower arm (left) and the linear acceleration along the same axis (acceleration without the gravity component that IMUs can derive with the help of gyro and magnetic field sensors).}
    \label{fig:linx_vs_ax_linx_2}
\end{figure}
\begin{figure}
    \centering
    \includegraphics[width=0.49\columnwidth]{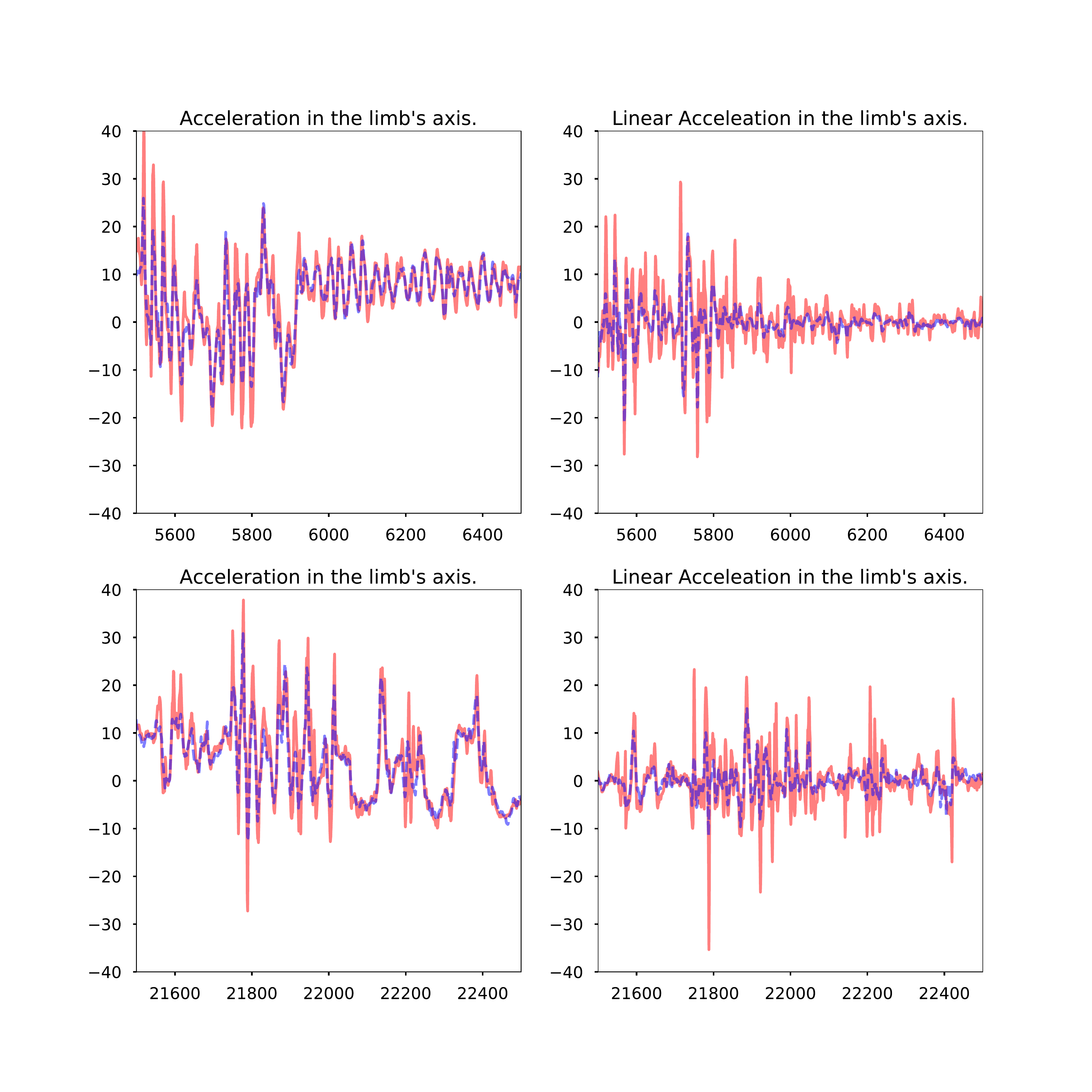}
    \includegraphics[width=0.49\columnwidth]{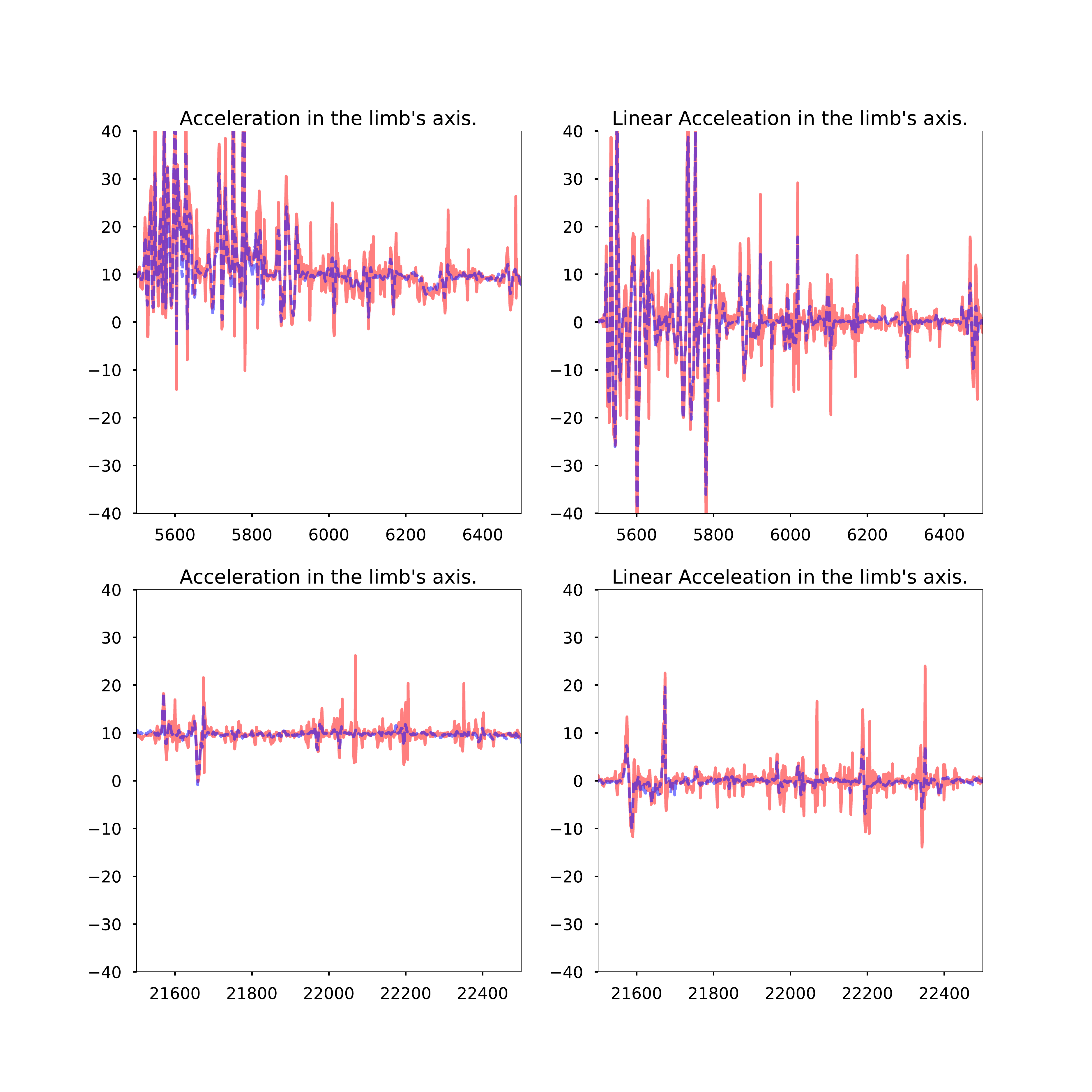}
    \caption{Real signals (red) versus simulated (blue) in the regression training set (generic motions) for acceleration and linear acceleration both in the limb's axis. All cases trained without neither filtering nor standard scaling.}
    \label{fig:lin_x_ax_signals}
\end{figure}

To probe further we have plotted signals from individual acceleration axis as both raw acceleration and linear acceleration in Figure \ref{fig:linx_vs_ax_linx_2} including a zoomed in version in Figure \ref{fig:lin_x_ax_signals}. 
The single axis acceleration signal is richer in features than both the linear acceleration on the same axis and the norm signals discussed above. This in itself is well known and not surprising. What is surprising is how well the simulated signals replicate the real ones given the issues described in Section \ref{sec:problem} ( e.g the difficulty of estimating the projection of gravity into individual axis). This applies to both the overall, broad structure of the signal (Figure \ref{fig:linx_vs_ax_linx_2}) and, in many  cases to the detailed strcutre. An example is illustrated in Figure \ref{fig:lin_x_ax_signals} on the left where  some very subtle features have been matched nearly perfectly. Clearly other examples exist as shown on the right of the Figure where there is much more difference between the features of the simulated and the real signal.  

\subsection{Effects of Pre-Processing}\label{sec:effect_prep}
\label{sec:effects_preprocessing}
From the above discussion it is apparent that the two main concerns are modelling high frequency components such as acceleration "ringing" caused by sudden forceful impacts and exact matching of the signal magnitude. This suggests the use of low pass filtering and scaling as obvious pre-processing approaches. Both are, as described in Section \ref{sec:preprocessing}, parts of our pipeline.   Note that we are talking about standard scaling when \textbf{learning the regression}, that is, with the mean and standard deviation computed on our dataset of generic motions being performed by other users. As the mean may be different in this dataset, we simply undo the standard scaling, which is fully reversible, when obtaining simulated signals. 
\subsubsection{Signal Level Effects of Filtering and Scaling}
Another pre-processing parameter within our pipeline are the sliding window sizes for training the regression.  We experimented with window sizes of $50$ and $16$ values, representing $1$ second and $0.32$ seconds, respectively. In our early experiments training the regression models it became clear that the smaller value was better, providing a smaller regression loss and better overall classification. Thus, for brevity, we report all results with the smaller window size. Regarding window step, we keep it at $1$
to increase the amount of training data.

\begin{figure}
    \centering
    \includegraphics[width=1.0\columnwidth]{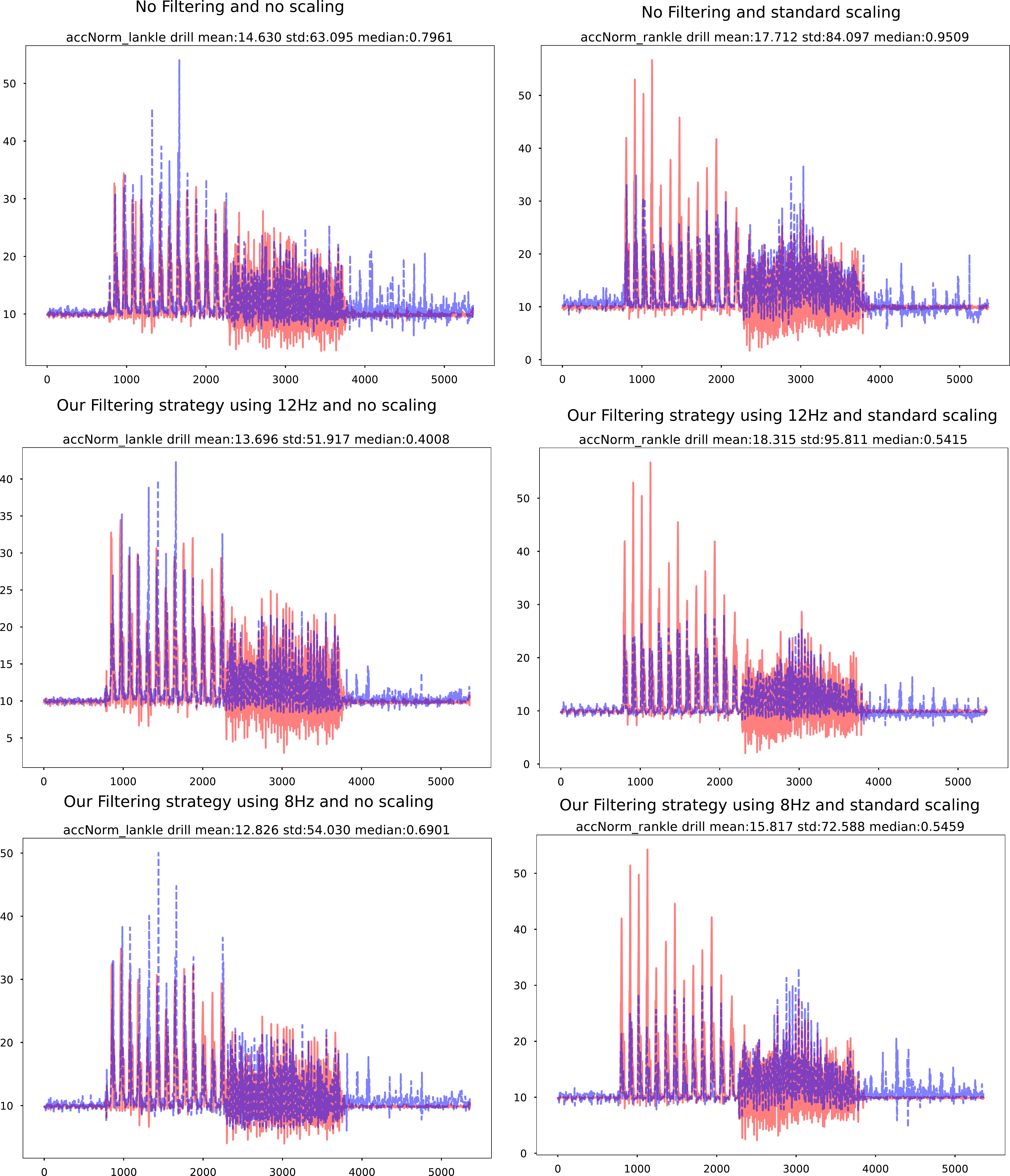}
    \caption{Examples signals generated by our method (blue) and their real counterparts (red) in the target dataset (fitness exercises) under different filter and scaling strategies.}
    \label{fig:fft_drill}
\end{figure}

\begin{figure}
    \centering
    \includegraphics[width=0.49\columnwidth]{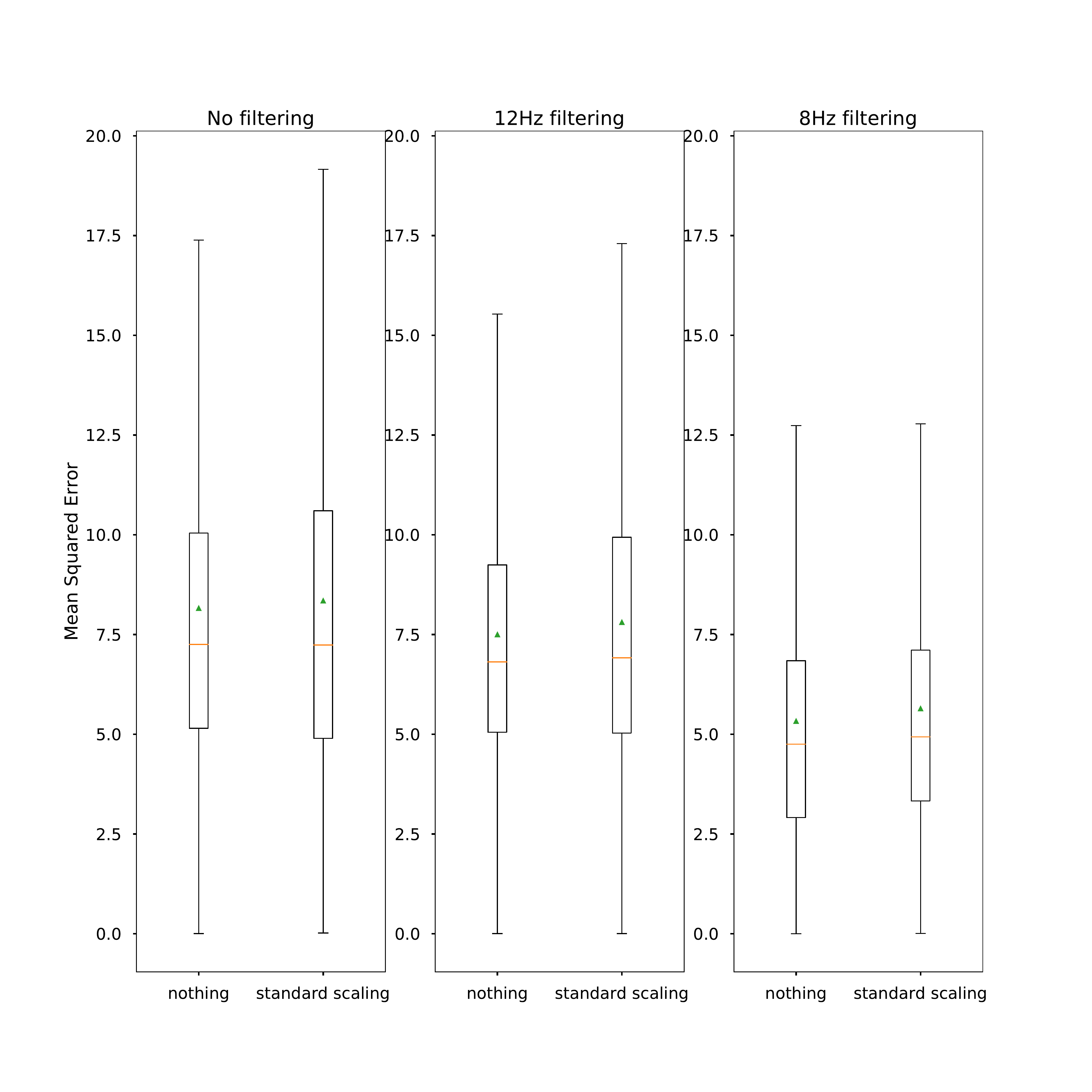}
    %\caption{ Loss for all windows using different training strategies for accelerometer norm. }
    %\label{fig:regr_error_seed_acc}
%\end{figure}
%\begin{figure}
    %\centering
    \includegraphics[width=0.49\columnwidth]{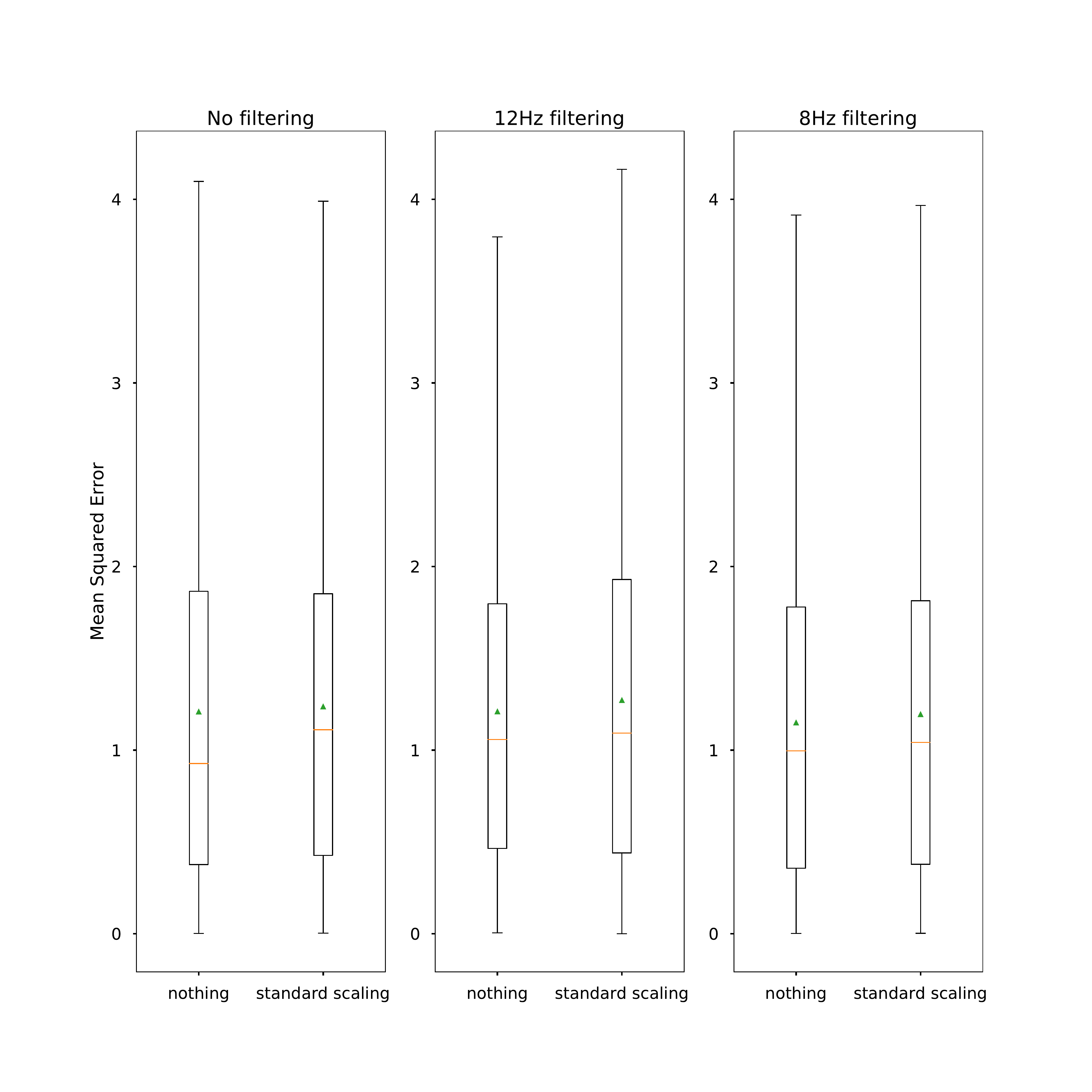}
    \caption{ Loss for all windows using different training strategies for acceleration (left) and gyroscope norm (right). }
    \label{fig:regr_error_seed_gyr}
    \label{fig:regr_error_seed_acc}
\end{figure}

In Figure \ref{fig:fft_drill} we illustrate the effect of frequency filtering and scaling on the performance of our regression model. We consider low pass filtering with 12Hz and 8Hz (given  the original rate of 50Hz from the video signals). The low pass filtering itself was done using a butterworth low pass filter of the $6$th order and standard scaling was done by removing the mean of the sensor channel in the training set and dividing by its standard deviation. Example comparisons of a signal with no filtering and an scaling on one hand and with 8Hz filtering and scaling on the other are also shown in Figure \ref{fig:moresignals} (left and right column respectively).
Key observations are:
\begin{enumerate}
    \item Overall there is no dramatic effect. In all cases the simulated signal follows the  structure of the real signal with reasonable accuracy while displaying the same types of problems.  
    \item Without scaling (left column on Figure \ref{fig:fft_drill}) there are some pronounced amplitude outliers in the high frequency peaks amplitudes (between sample 1000 and 2000) that are not influenced by the frequency filtering.  These disappear in the scaled versions (right column), however at the cost of significant  underestimation of the amplitude in the same area and overestimation around 3000. 
    \item In Figure  \ref{fig:fft_drill} frequency filtering has little effect, except for reducing the underestimation of the below $1g$ components around sample 3000 in the no scaling case (which overall seems to be the best). 
\end{enumerate}
In  summary while both filtering and scaling do have benefits in some situations, the qualitative signal examination described above does not indicate it to be decisive or obvious. To get a more quantitative idea Figure \ref{fig:regr_error_seed_gyr} shows the Mean Squared Error for computed over all our samples for acceleration and gyroscope for different filter values. For acceleration the error is indeed smaller for 8Hz (but not by much). For the gyroscope signal there is no significant difference. 

\subsubsection{Classification Level Effects of Filtering and Scaling}
\label{sec:class_level_filter}
\begin{figure}
    \centering
    % \begin{subfigure}{1.0\columnwidth}
        \centering
        \includegraphics[width=0.95\columnwidth]{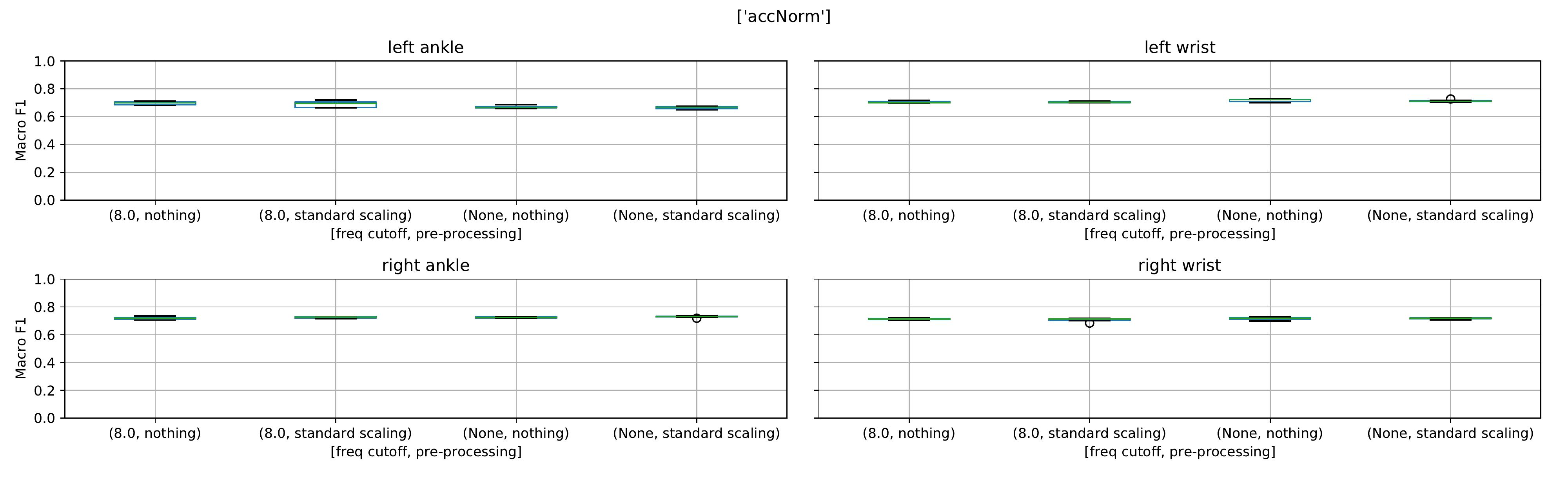}
        \centering
        \includegraphics[width=0.95\columnwidth]{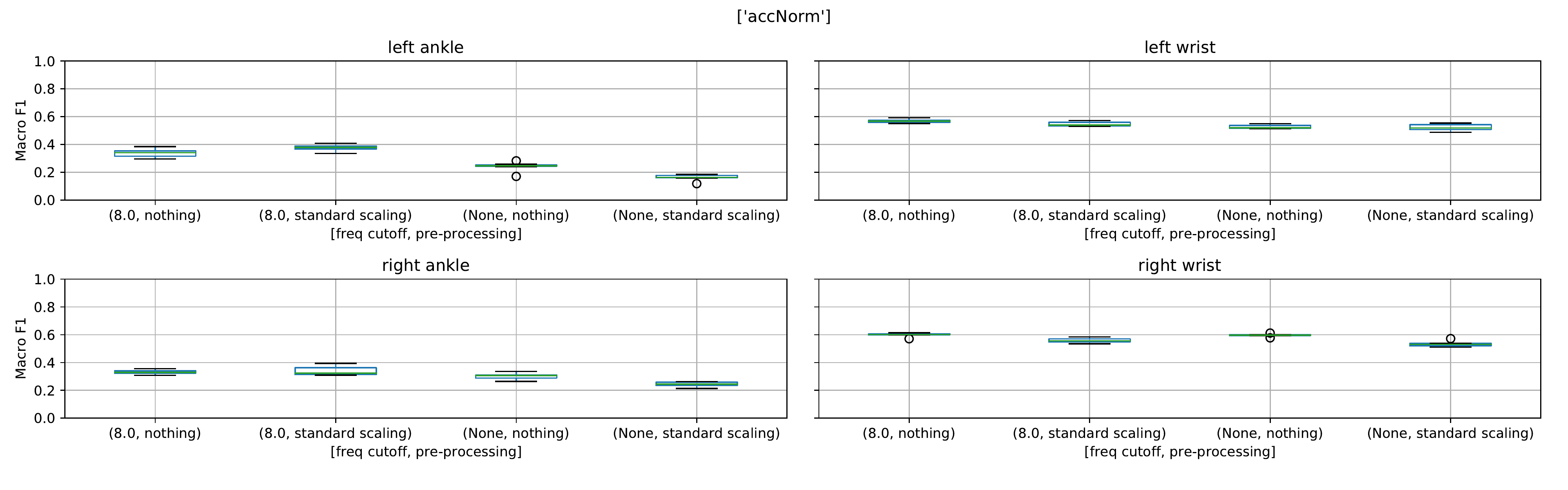}
        % \caption{simulated acceleration norm}
    % \end{subfigure}%
    \caption{ Comparison of classification performance for models trained with real (upper) or simulated data (down) under different pre-processing techniques using the acceleration norm. Results are shown for each different sensor placement separately. }
    \label{fig:hyp_diff_params_newest}
\end{figure}

As a final evaluation step we did an evaluation regarding the impact of filtering and scaling on the classification performance. To this end we need to consider not only how the pre-processing impacts the comparison between the classification within simulated sensor data but also how it impacts the performance of the absolute recognition rates of the real sensor data based models. It makes little sense to apply a pre-processing methods that makes the results of the simulated signals based model equal to that of a real signals based one but at the cost of making both much worse.
In Figure \ref{fig:hyp_diff_params_newest} we thus consider the effect of 8Hz filtering and scaling in different combinations on the recognition  for different placements (wrist and ankle) of the sensor for the real and the simulated data.  The training was done on all the data designated for training and the testing on all testing data  as described in Section \ref{sec_eval_proc}. We can see that the recognition rates on the real data show little sensitivity to the filtering and scaling. The performance on the simulated data is also fairly invariant for the wrist sensor placement while for the ankle placement frequency filtering does indeed make a significant difference. However, it must also be noted that for the ankle placement the recognition rate with the simulated data is very poor to start with (around 40\%). This will be discussed in the next section in the context of the impact of sensor selection. 
%----------

%\subsubsection{Different Types of Training Data and Sensor Sets}
\begin{figure}
    \centering
    \includegraphics[width=0.49\columnwidth]{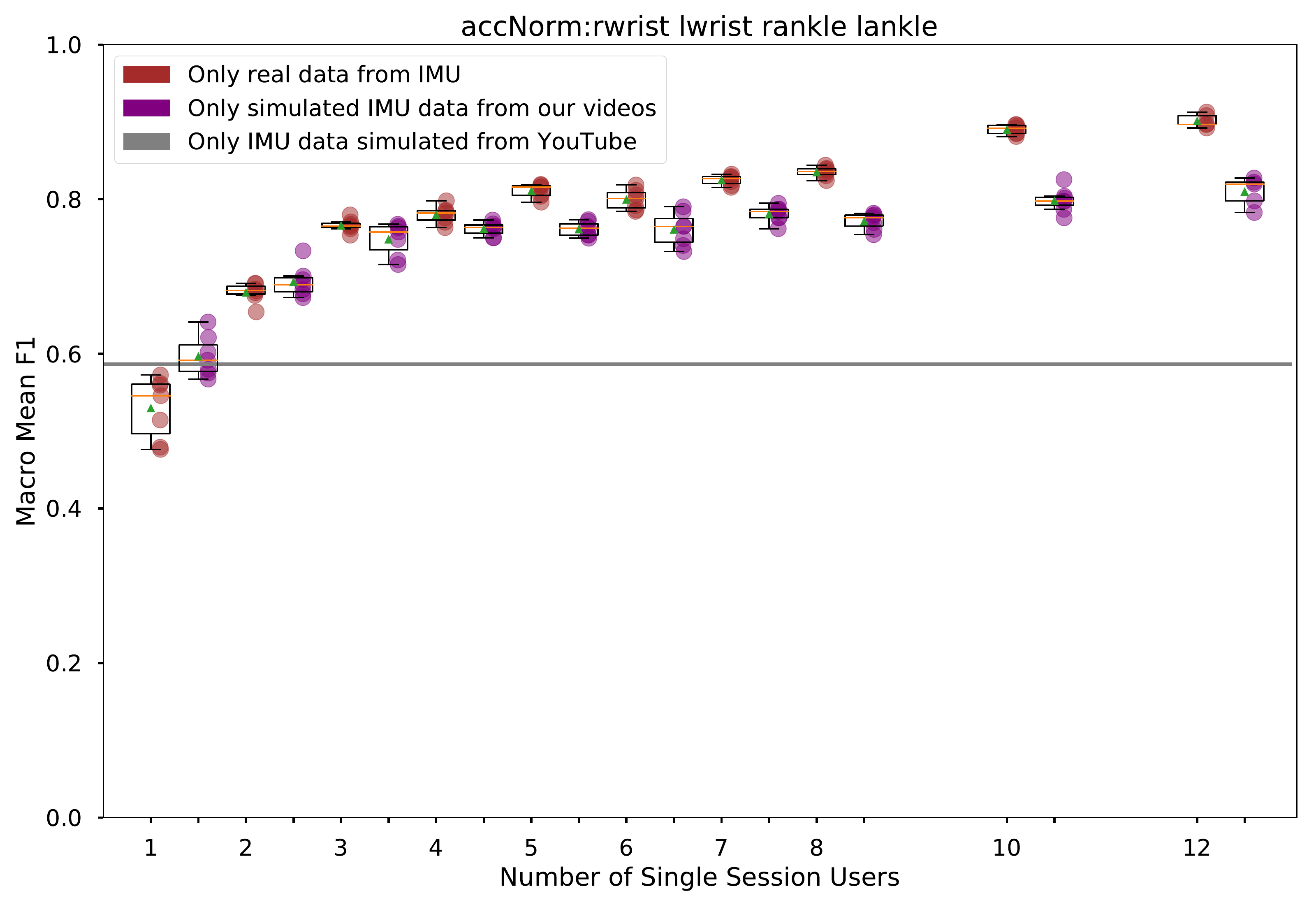}
    % \label{fig:results_acc}
    \centering
    \includegraphics[width=0.49\columnwidth]{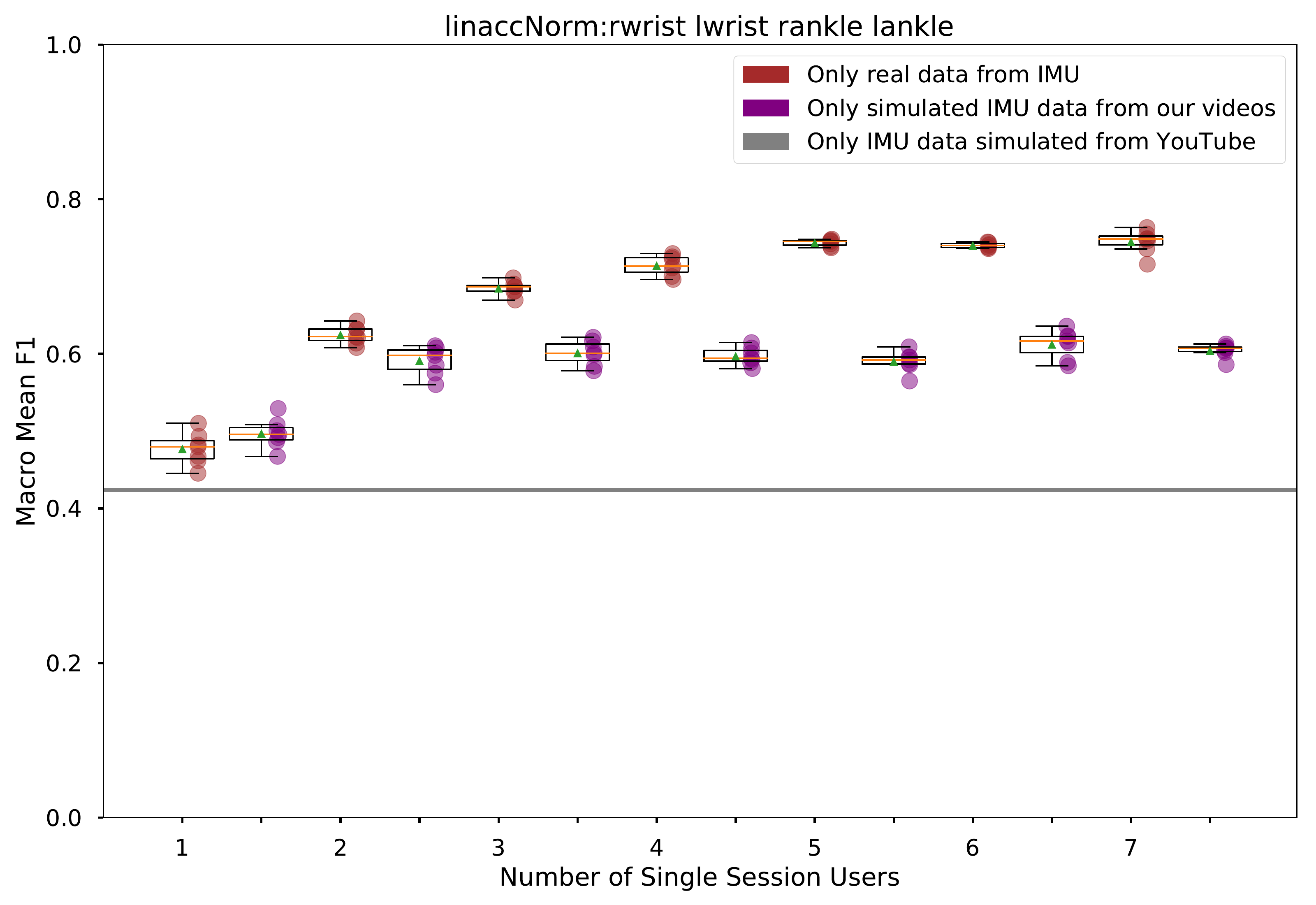}
     %\label{fig:results_linacc}
%\end{figure}
%\begin{figure}
    \centering
    \includegraphics[width=0.49\columnwidth]{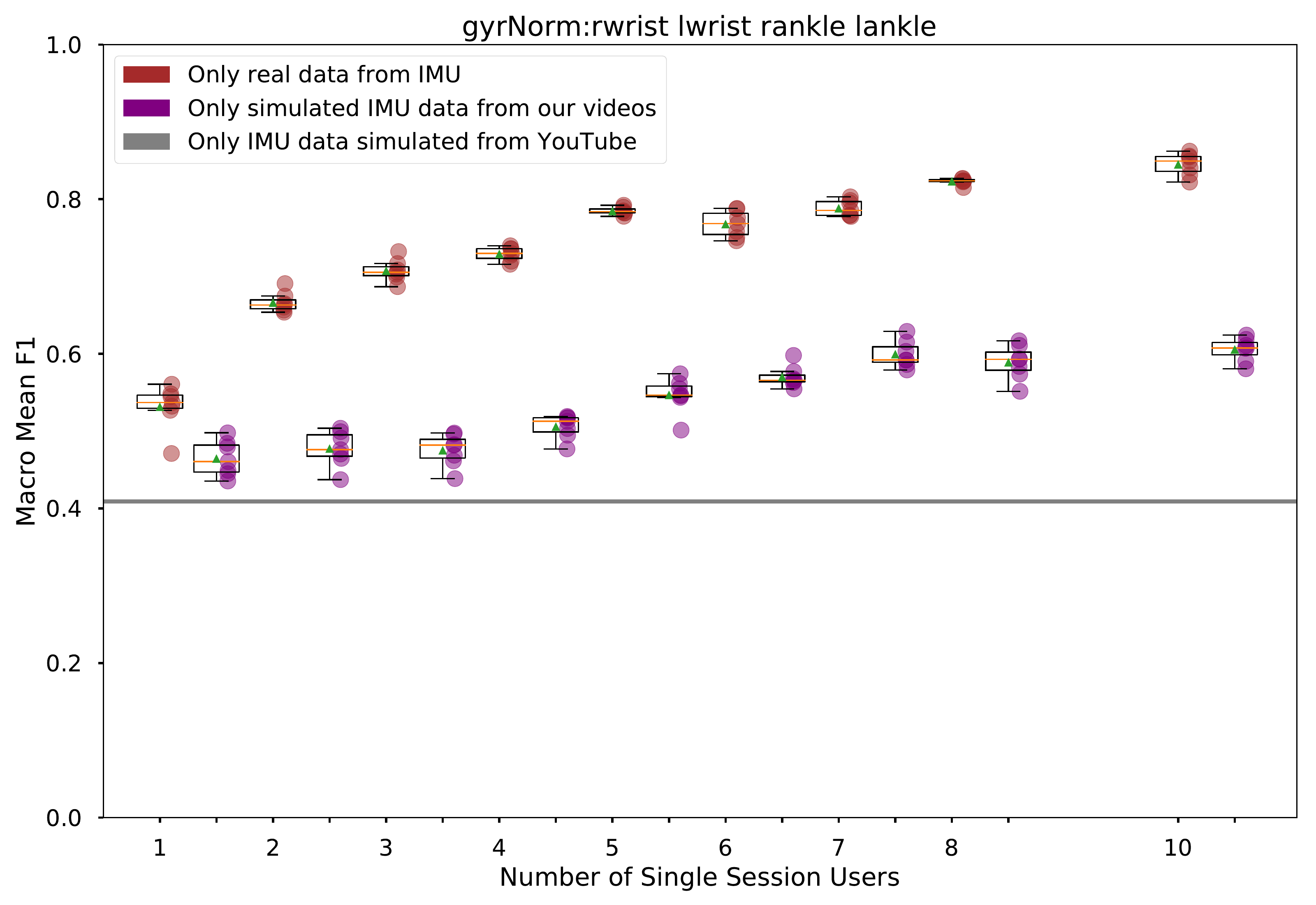}
%    \caption{Results when using the gyroscope norm. }
    %\label{fig:results_gyr}
%\end{figure}
%\begin{figure}
    \centering
    \includegraphics[width=0.49\columnwidth]{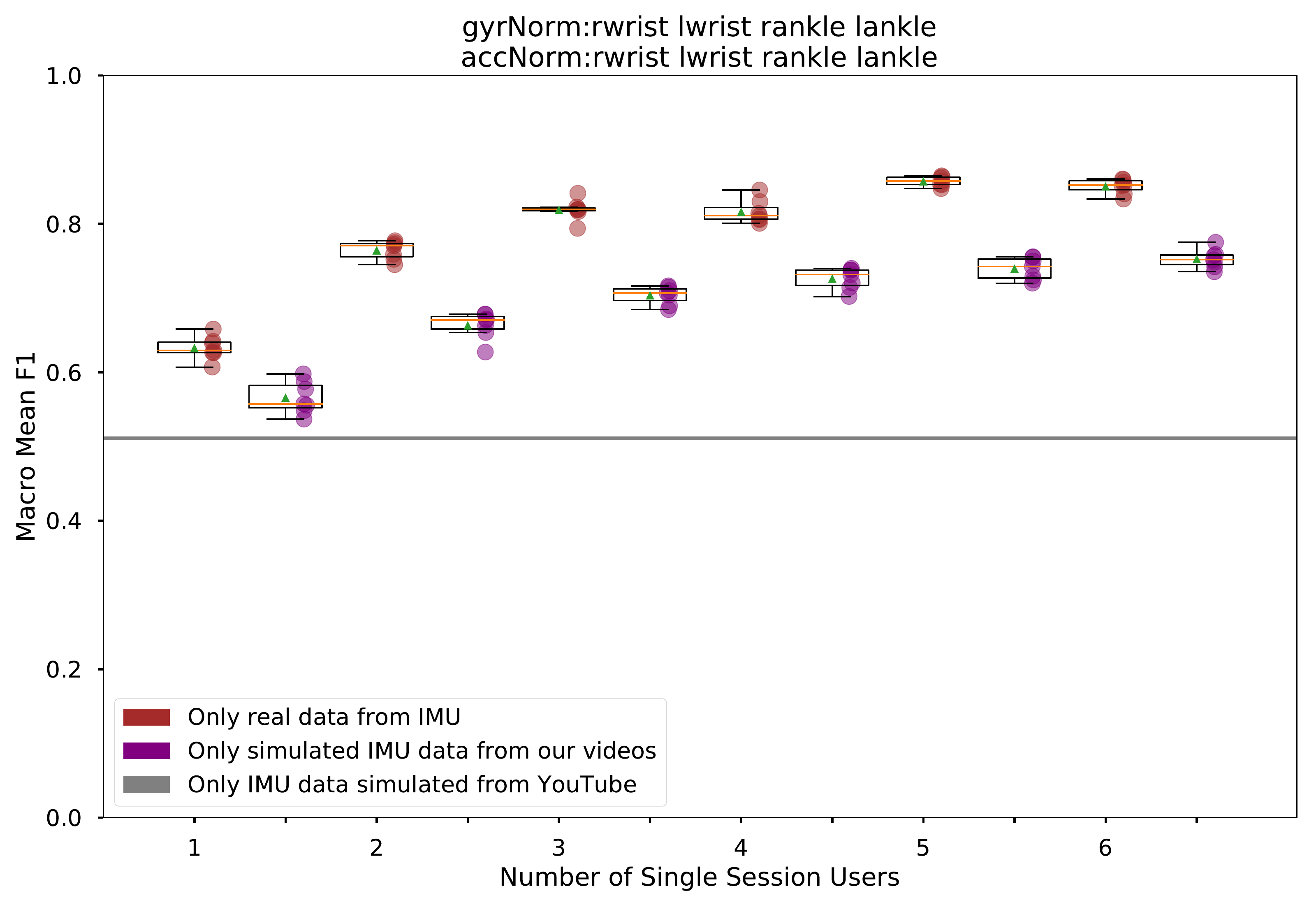}
    \caption{Results when using the accelerometer and gyro norm and the combination thereof.}
    \label{fig:results_both}
\end{figure}

\subsection{Evaluation of Classification Performance Using the Simulated Signals}\label{sec:eval_perf_all}
\begin{figure}
    \centering
    \includegraphics[width=0.49\columnwidth]{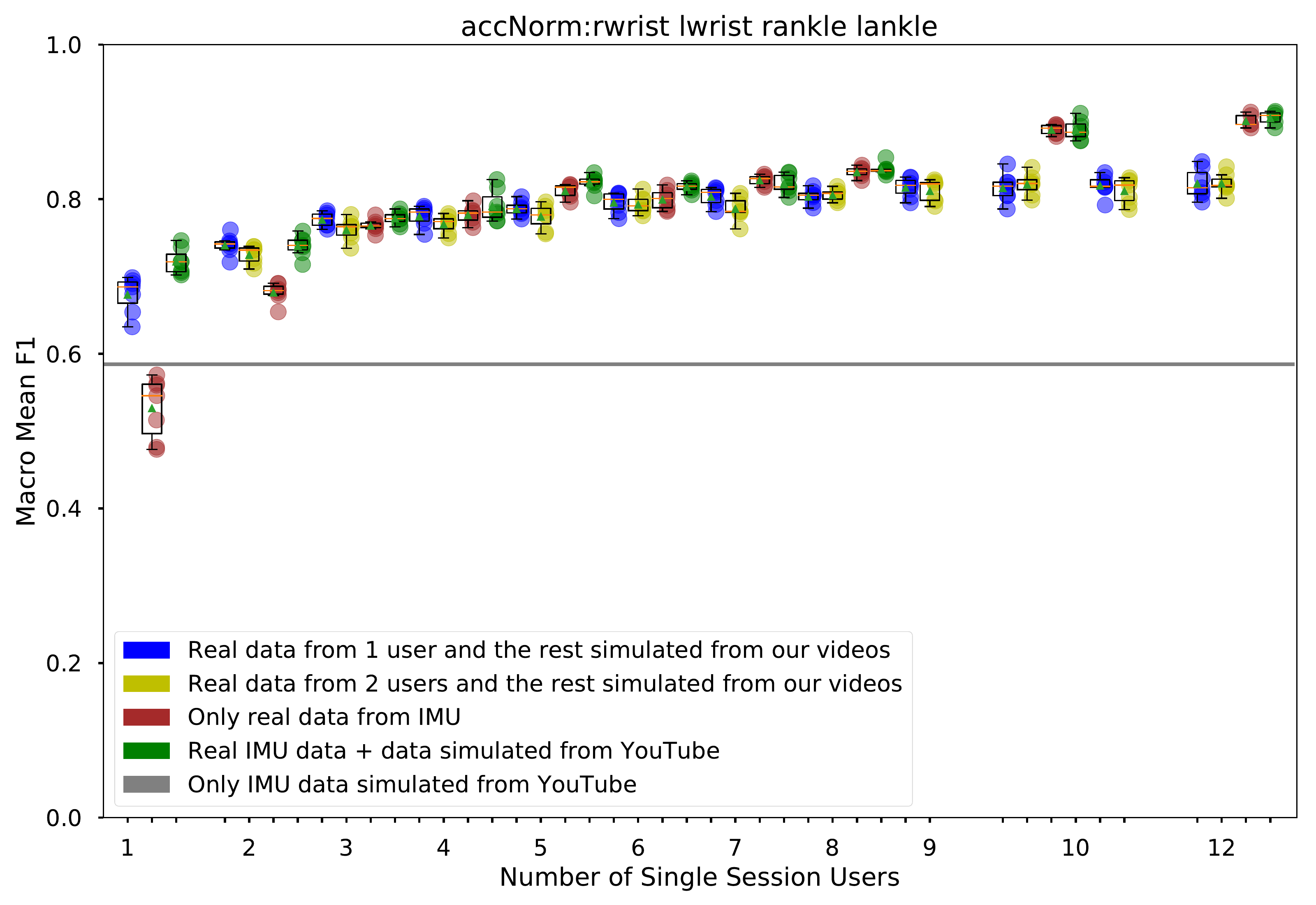}
    %\caption{Results when using the accelerometer norm and adding simulated data to the real one. }
    % \label{fig:results_acc}
%\end{figure}
%\begin{figure}
    \centering
    \includegraphics[width=0.49\columnwidth]{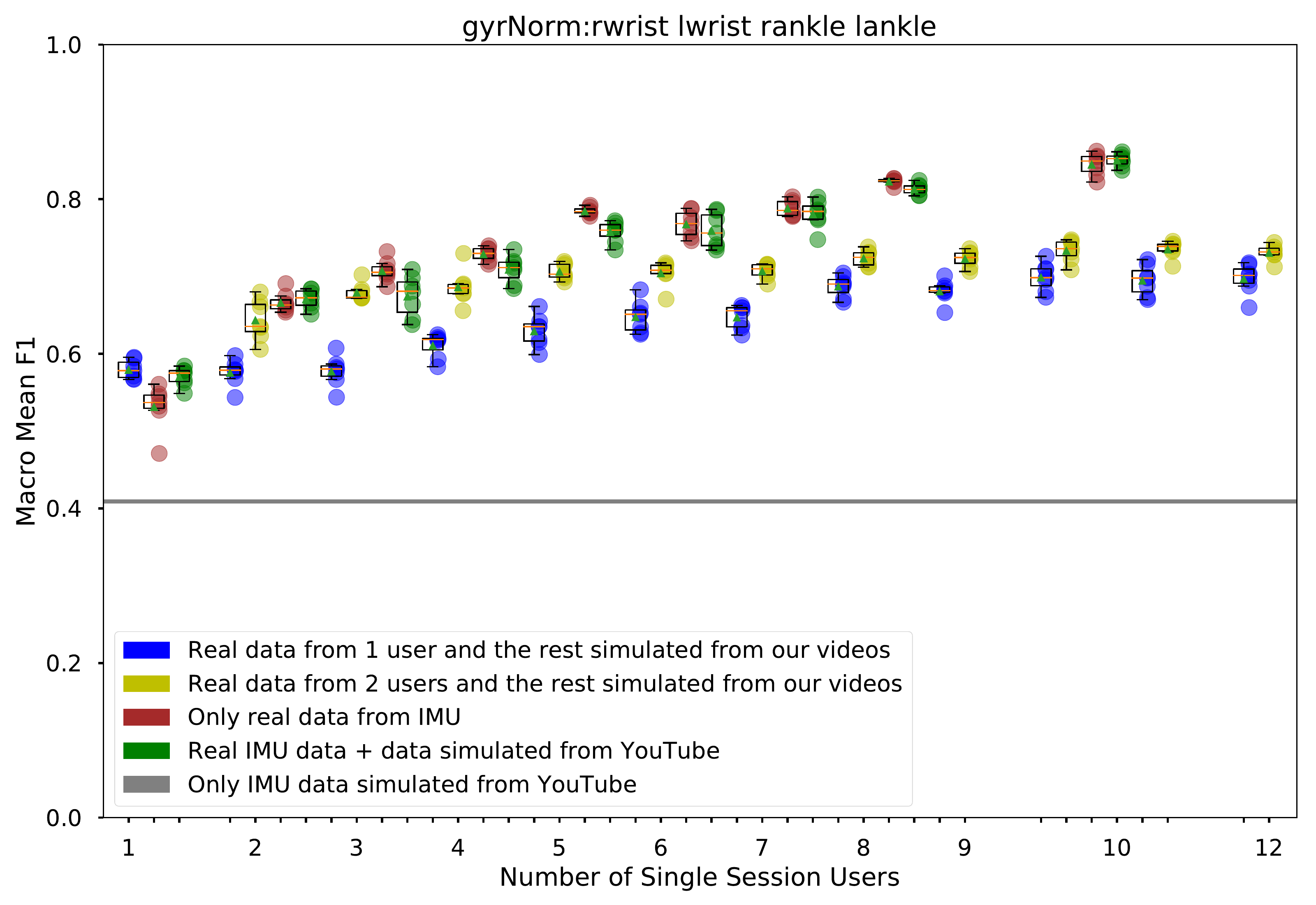}
    %\caption{Results when using the gyroscope norm and adding simulated data to the real one. }
    % \label{fig:results_gyr_mix}
%\end{figure}
%\begin{figure}
    \centering
    \includegraphics[width=0.49\columnwidth]{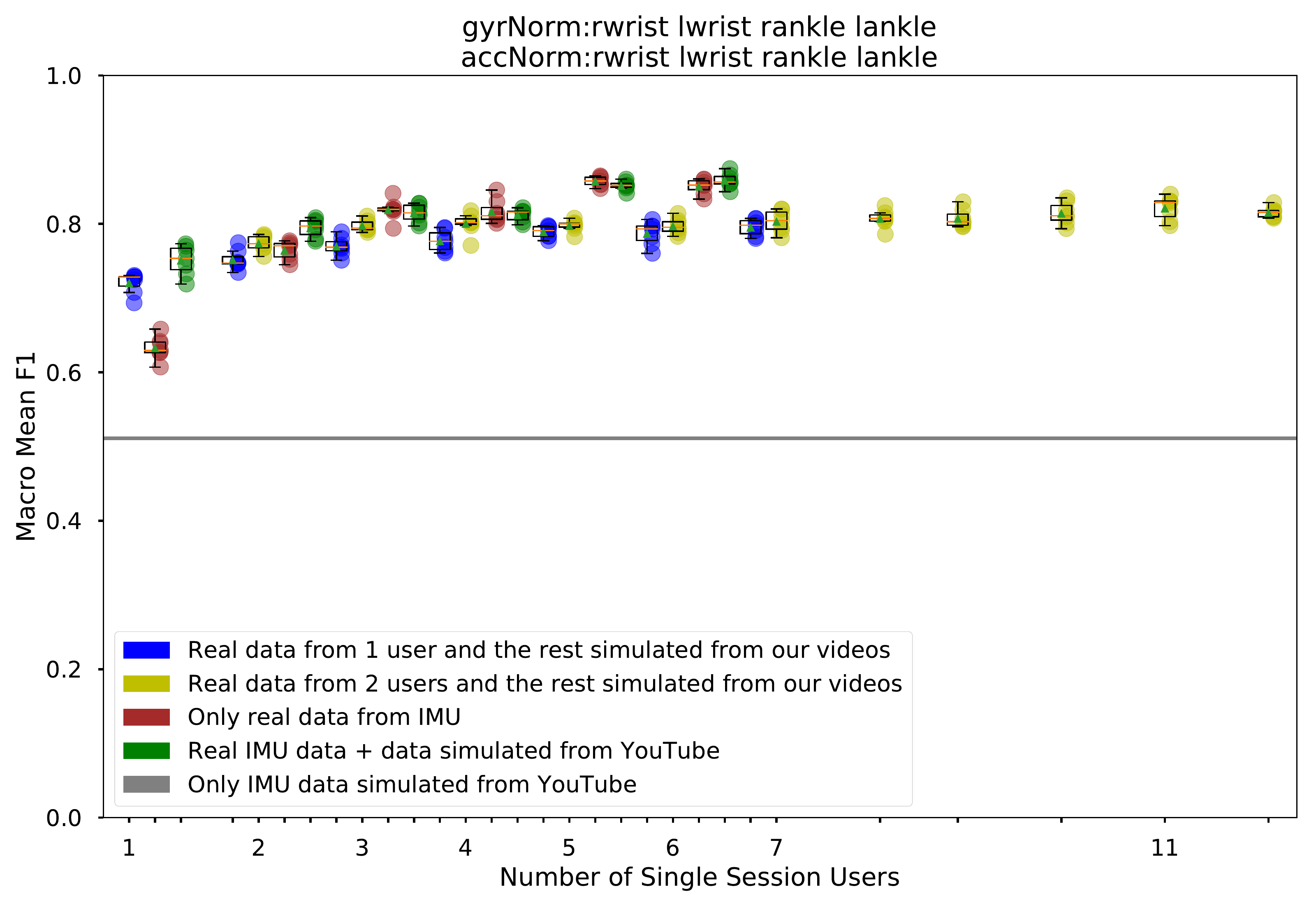}
    \caption{Results when using the gyroscope and accelerometer norms and adding simulated data to the real one. }
    \label{fig:results_both_mix}
\end{figure}

\begin{figure}
    \centering
    \includegraphics[width=0.49\columnwidth]{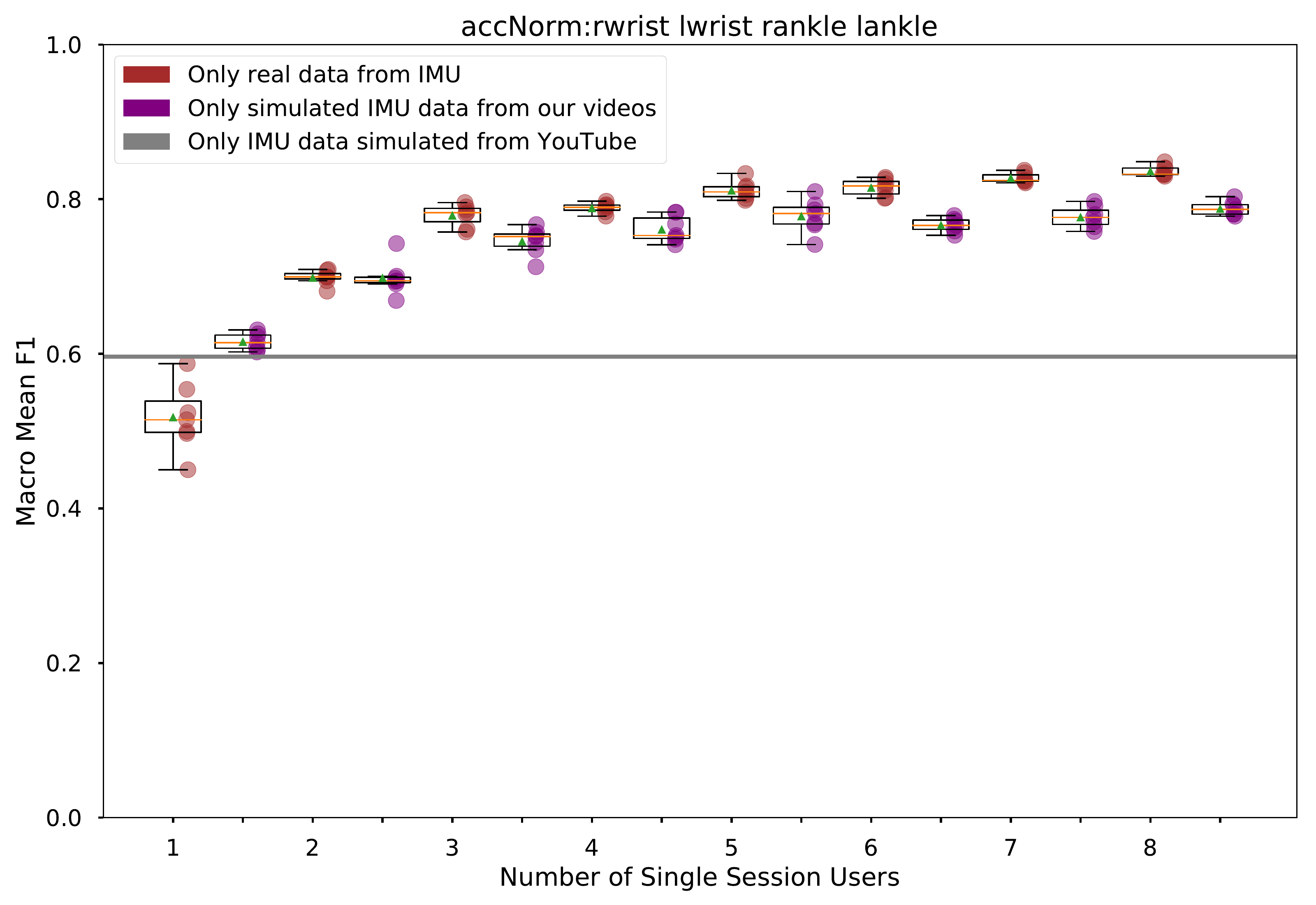}
   % \caption{Results when using the accelerometer norm and also 8Hz filtering. }
    \label{fig:results_acc_8hz}
%\end{figure}
%\begin{figure}
    \centering
    \includegraphics[width=0.49\columnwidth]{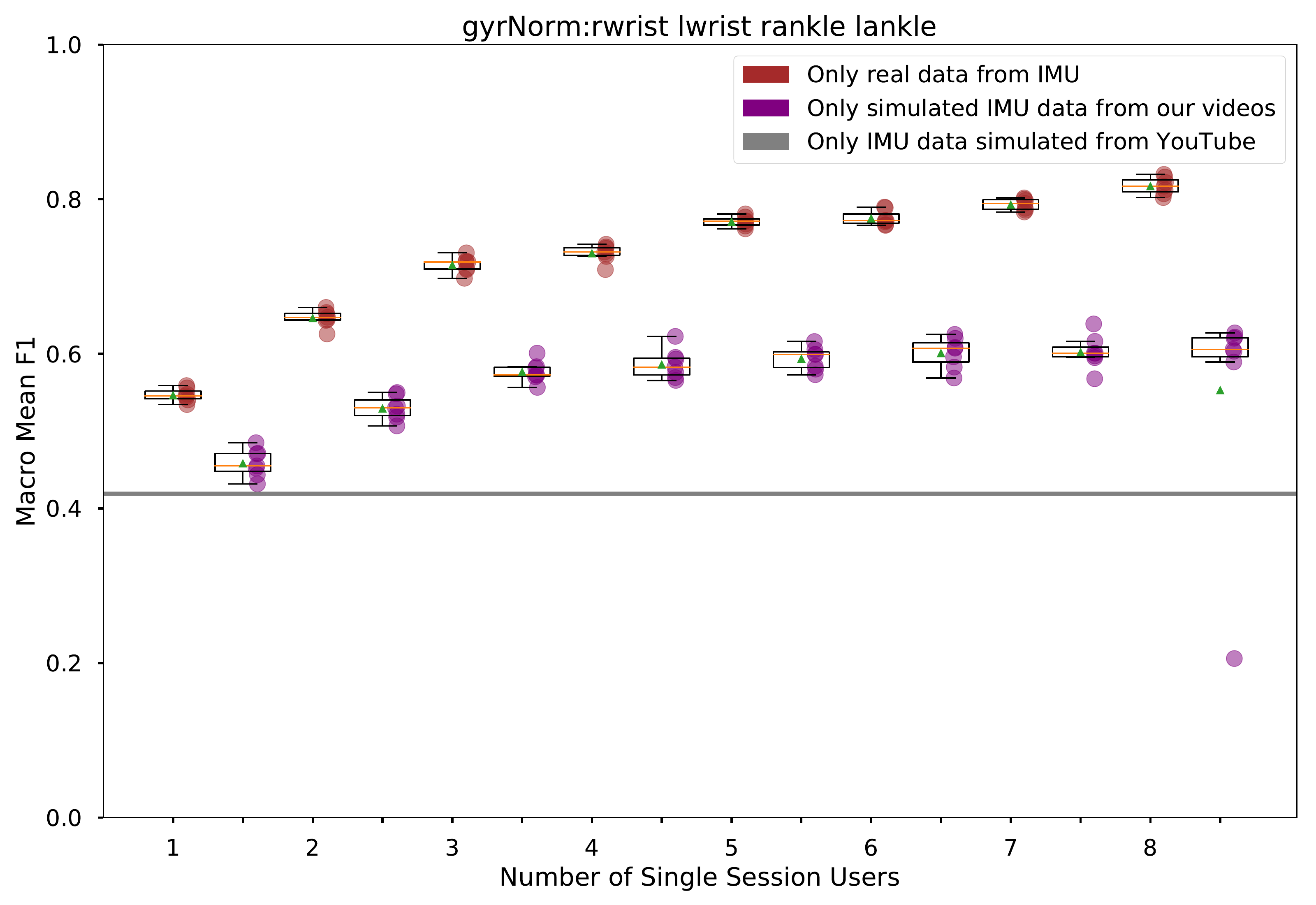}
    %\caption{Results when using the gyroscope norm and also 8Hz filtering. }
    \label{fig:results_gyr_8hz}
%\end{figure}
%\begin{figure}
    \centering
    \includegraphics[width=0.49\columnwidth]{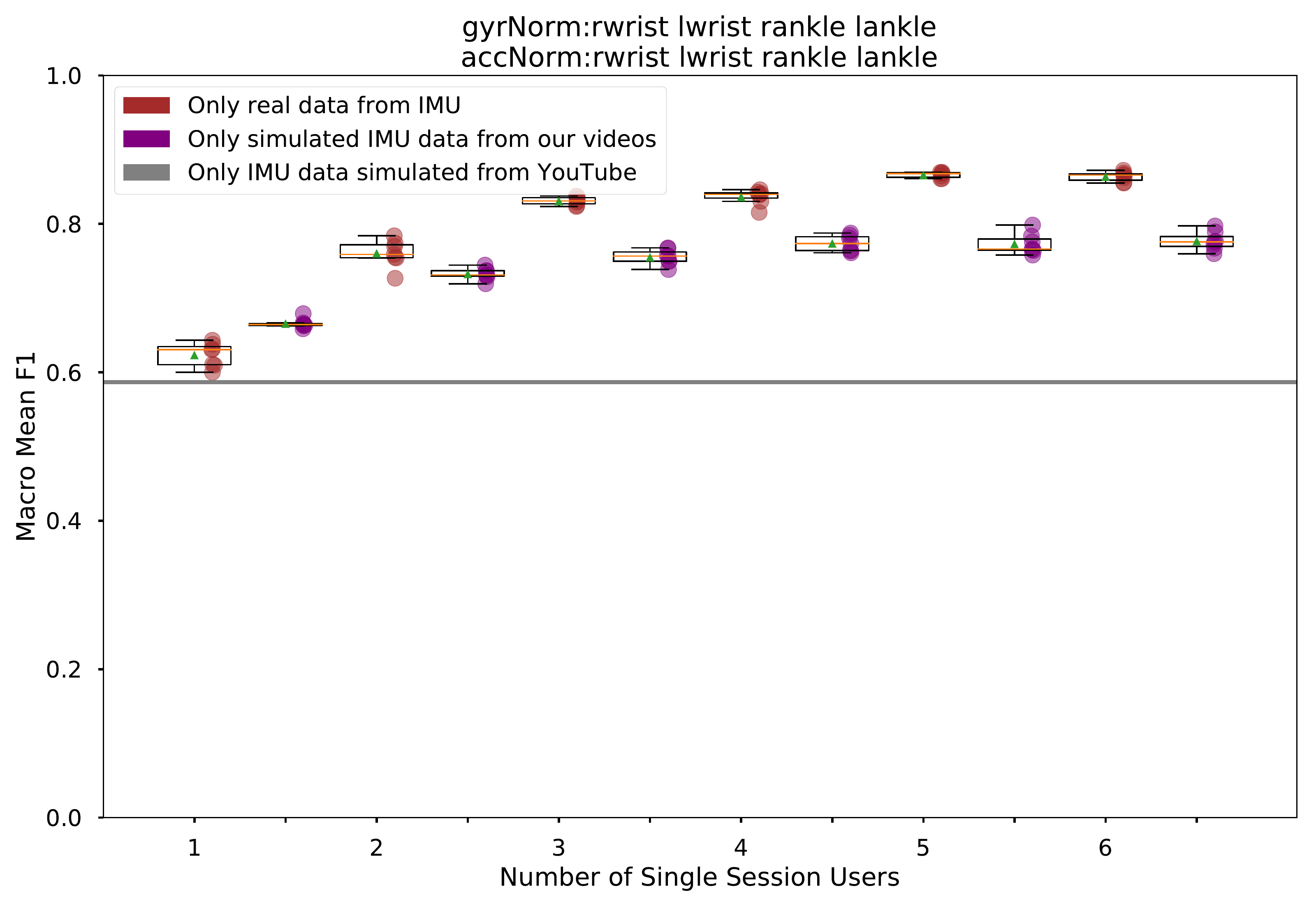}
    \caption{Results when using the gyroscope and accelerometer norms and also 8Hz filtering. }
    \label{fig:results_both_8hz}
\end{figure}

The discussion in the previous section has shown that, mostly on a qualitative level, our method generates signals that replicate a significant portion of the respective real sensor output within our application domain. We now proceed to quantitatively evaluate how well such simulated sensor signals are suited for training activity recognition that will later be applied to real sensor signals. The evaluation procedure has already been described in Section \ref{sec_eval_proc}. Essentially we have recorded a dataset where for each activity and user we have both a video signal and sensor data. In addition we have collected some YoutTube Videos for the same/similar activities. We then train classifiers using the real sensor data, simulated sensor data from the videos that we have recorded, simulated sensor data from the YouTube videos and various combinations thereof. The classifier trained on real sensor data is the  to which we compare the performance of the different combinations to evaluate the usefulness of our approach. Given the discussion in the previous section we focus on using  raw data without filtering and scaling with some examples of the impact of 8Hz filtering. The results are shown in Figures, \ref{fig:results_both}, \ref{fig:results_both_8hz} and \ref{fig:results_both_mix}. In each Figure we plot the recognition rate of each classifier vs. the number of users in the training set.  The only exception is the case of sensor data generated from YouTube videos where we had different users perform different exercises from the set which means that the notion of "number of users" for the whole  dataset makes no sense (see Table \ref{tbl:yt}). Instead we have generated training data from the entire 14min of the YouTube data that we have harvested and plotted it for comparison as a constant in the graphs. 

\begin{enumerate}
    \item {\bf Baseline.} The recognition rates on the real data reach up to 90\% which, given the fact that our aim was not to optimize a recognition system for a specific application, is an acceptable baseline to evaluate the  performance of the simulated sensor data.
    \item {\bf Overall performance on simulated sensor data.}  Systems trained on the simulated data reach up to 80\% recognition rate, which is 10\% below the  results (see e.g. Figure \ref{fig:results_both} left where the acceleration norm based recognition is among the best performing variants overall). In general as the number of users is small the results for  real and simulated data are very similar with the performance of  the real data. The performance of the real data then more quickly improves with the amount of training data. This is to be expected as with a very small number of users (equal a small amount of training data) the recognition rate tends to be poor and  limiting factor for the performance is the lack of diversity in the data and not the data quality. As the amount of data increases the quality becomes the limiting factor, which is better for the real sensor data. 
    \item {\bf Performance on YouTube data.}  The performance of the system based on sensor data simulated from YouTube videos is in most cases slightly below the performance of the baseline on a single or two users. Given that the total amount of YouTube data (around 14 min) is in the order of magnitude of the length of the data from a single user this is not surprising. 
    \item {\bf Compensating deficiencies of simulated data through training set size.}  There is strong indication that deficiencies of the simulated sensor data can be compensated by the amount of such data. Thus in most cases (see discussion of different sensor combinations later on) systems based on simulated sensor data can match the performance of the  real sensor data based system trained on about   half as much data. Again looking at the acceleration norm from all sensors case in Figure \ref{fig:results_both} left top we see that the Mean F1 score for simulated data with 12 users is about the same as the score for 6 users with real data. Furthermore, while we see a slower growth of the F1 score with the number of users in the training set for the case of simulated data then for real data, in most cases a growth exists. This is in line with the vision outlined on the introduction  that aims to leverage the tremendous amount of available online video data to create robust models for wearable sensors based systems  from simulated training data based on those videos. 
    \item {\bf Effects of model fine tuning using real sensor data.} Another way to improve the performance of the simulated sensor data based systems is to use small amounts of real sensor data to fine  tune it. The idea is that collecting a small amount of real data if nearly always possible, it is the bulk of a large training data set that is difficult to collect and that we want to get from existing videos. The results of this strategy are illustrated in Figure \ref{fig:results_both_mix}. We compare the baseline to the performance of a system trained on simulated data to which 1 (blue points) or 2 (yellow points) users with real data have been added.  We can see that the simulated data is now much closer to the real one. In the top left graph showing the acceleration norm results it is virtually identical up to 10 users when the real data starts to overtake the simulated one.  The effect is even more significant for the gyro norm based recognition (see Figure \ref{fig:results_both_mix} right top)  where the purely simulated  signals based model trailed the real signals based ones by 30\% and more (see Figure  \ref{fig:results_both} left bottom) and are now not worse than about 10\% worse (and up to 6 users nearly identical).  
    \item {\bf Effects of adding simulated data to small amounts of real data.} Another situation where a combination of real and simulated signals may be useful is when we have only a small amount of sensor based training data with no easy  possibility  of collecting more. We then try to "top up" our  training set with some simulated data from video. The effect of such a strategy is illustrated by the green points (real IMU data with YouTube) in Figure \ref{fig:results_both_mix} where we combine the real sensor data with the YouTube data increasing the number of real data over the x-Axis.  We can see that for small user numbers the strategy indeed helps. 
    \item {\bf Performance of different sensors and sensor  combinations.} Most of the discussion so far has been done with respect to the acceleration norm (Figures \ref{fig:results_both} and  \ref{fig:results_both_mix}) applied to both wrists and both ankles together  which has been the most effective modality with respect to the recognition performance both in the real sensor signal and with respect to how well the simulated signal can approximate it.  Given the type of activities that we use as our test set this is not surprising. The activities are characterized by periodic, mostly strong motions of various limbs. In most cases the relative temporal pattern and relative intensity of the motions is a characteristic feature.  At the same time the acceleration norm (in particular  when looking at the relative intensity and temporal patterns) is fairly robust against variations and noise. This also means that imperfections caused by the simulation of the signal from the  video data using our regression model can be well tolerated. 
    
    Further sensor modalities and their combinations that we investigated are:
    \begin{enumerate}
        \item {\em Acceleration norm  vs. Gyro norm vs. combination.} In Figure \ref{fig:results_both} we have in addition to the acceleration norm the linear acceleration norm the gyroscope norm and the combination of gyroscope and acceleration norms (in all cases for all the 4 sensor placements). The first thing we see is that that the linear acceleration leads to a poorer overall performance while having little impact on the relative performance of the simulated data. The former  is not surprising since the gravity component (meaning vertical orientation) is a very important piece of information.  The later is contrary to what we might have expected since our model was not tuned to capture gravity effects (as discussed in Section \ref{sec:problem}).  Apparently it was still able to capture sufficient amount of orientation related information. 
        Second, we see that the gyro signal has significantly worse relative performance for the simulated signal. Given the limitations on  recognizing rotations around certain axis (in particular the limb axis)  from the video signal discussed in Section \ref{sec:problem} this was to be expected. The very poor performance of the simulated gyro signal also drags down the combined acceleration/gyro norm performance shown in the bottom right part of  Figure  \ref{fig:results_both}.
        \item {\em Acceleration and linear acceleration on the limb's axis.} Further looking at the impact of linear acceleration  Figure \ref{fig:lin_x_ax_signals} shows the results for raw acceleration and linear acceleration along the limb axis. The idea is that the acceleration along the axis is independent of the rotation around the axis, which, as already explained is hard to exactly capture from video. It can be seen that both have slightly lesser performance than the acceleration norm (which is not surprising given that the acceleration norm is a good discriminator for our set of activities as described above). Within expected statistical bounds the difference between the simulated and the real sensor signal is about the same for both the raw and the linear acceleration. 
        \item {\em Different sensor placements.} In  Section \ref{sec:class_level_filter} we have already considered  the use of sensors from only a wrist or only the ankle (with respect to acceleration norm) as shown in Figure  \ref{fig:hyp_diff_params_newest}.  The performance for the real sensor data on both the wrist and the ankle  and the simulated sensor data on the wrist was with around 10-20\% less then for then for the combination which is not a surprising result (both legs and arms are relevant for our test activities).  What may be  surprising at first is the fact that the performance for the ankle using simulated data is at half of the real data (30-40\%). The explanation lies on the fact that the foot motions involve a lot of hard ground impacts that lead to "ringing" that has already often been mentioned as something that the simulated data can not replicate (which is why 8Hz filtering helps a lot here). In addition vertical orientation plays a bigger role then for the wrists and there are many motions "to the front" (towards the camera) which are more difficult to resolve due to the viewing angle. 
    \end{enumerate}
    \item {\bf Effects of 8Hz Filtering.} The effect of various pre-processing strategies has already been discussed in Section \ref{sec:effects_preprocessing} establishing that while filtering and to some degree scaling did seem be beneficial is some cases overall the effect was no overwhelming. Given that these effects may be strongly application specific. This is why we have decided to do most of the analysis in this section on the unfiltered, not scaled signal. However, for comparison Figure \ref{fig:results_both_8hz} includes the same evaluation done in Figure \ref{fig:results_both} with 8Hz filtering. We can see that while for the gyro norm (which was the poorest one to start with) the filtering does indeed improve the the relative performance of the simulate signal based model, it has little effect otherwise. 
\end{enumerate}

\begin{figure}
    \centering
    \includegraphics[width=0.49\columnwidth]{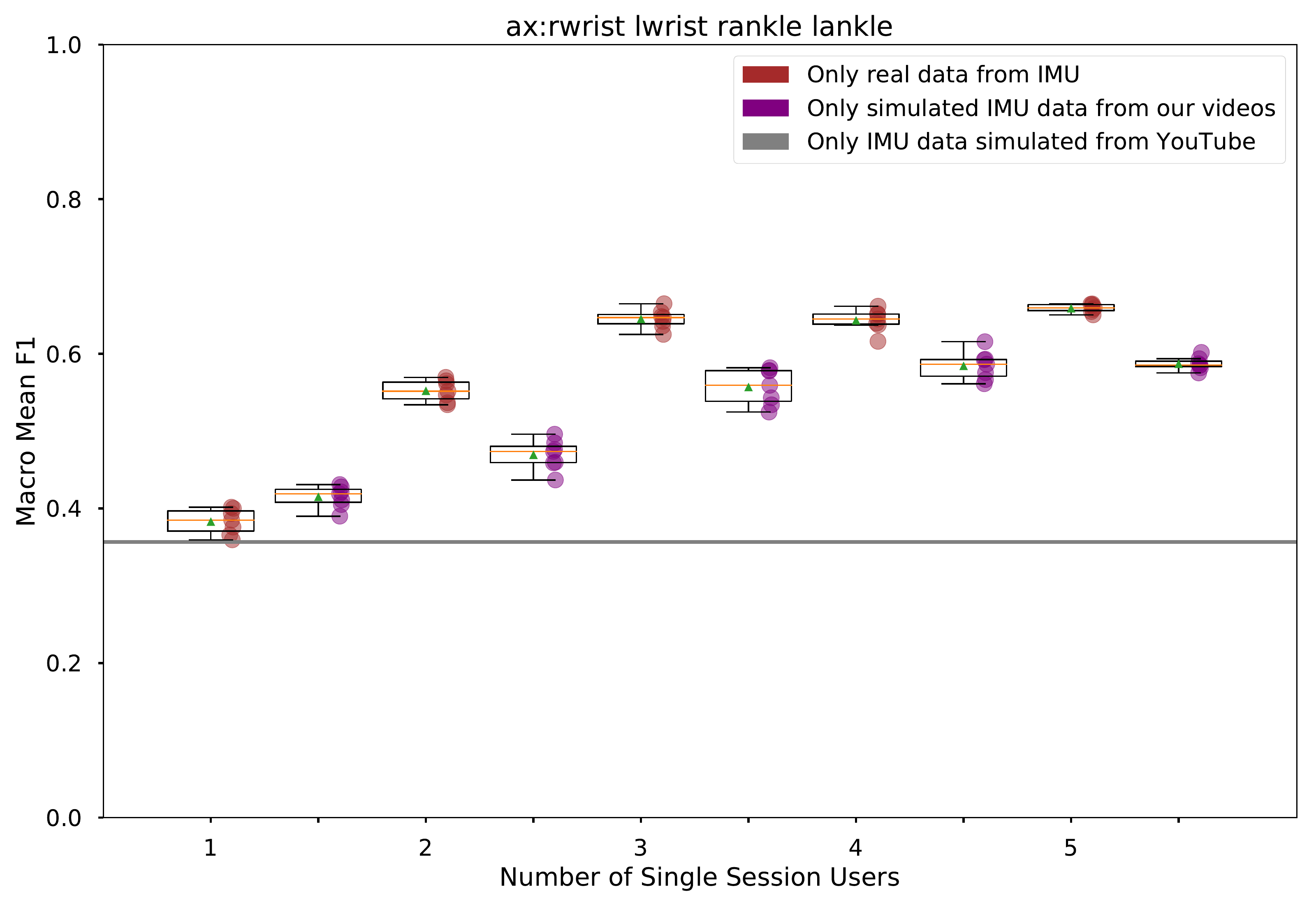}
    \centering
    \includegraphics[width=0.49\columnwidth]{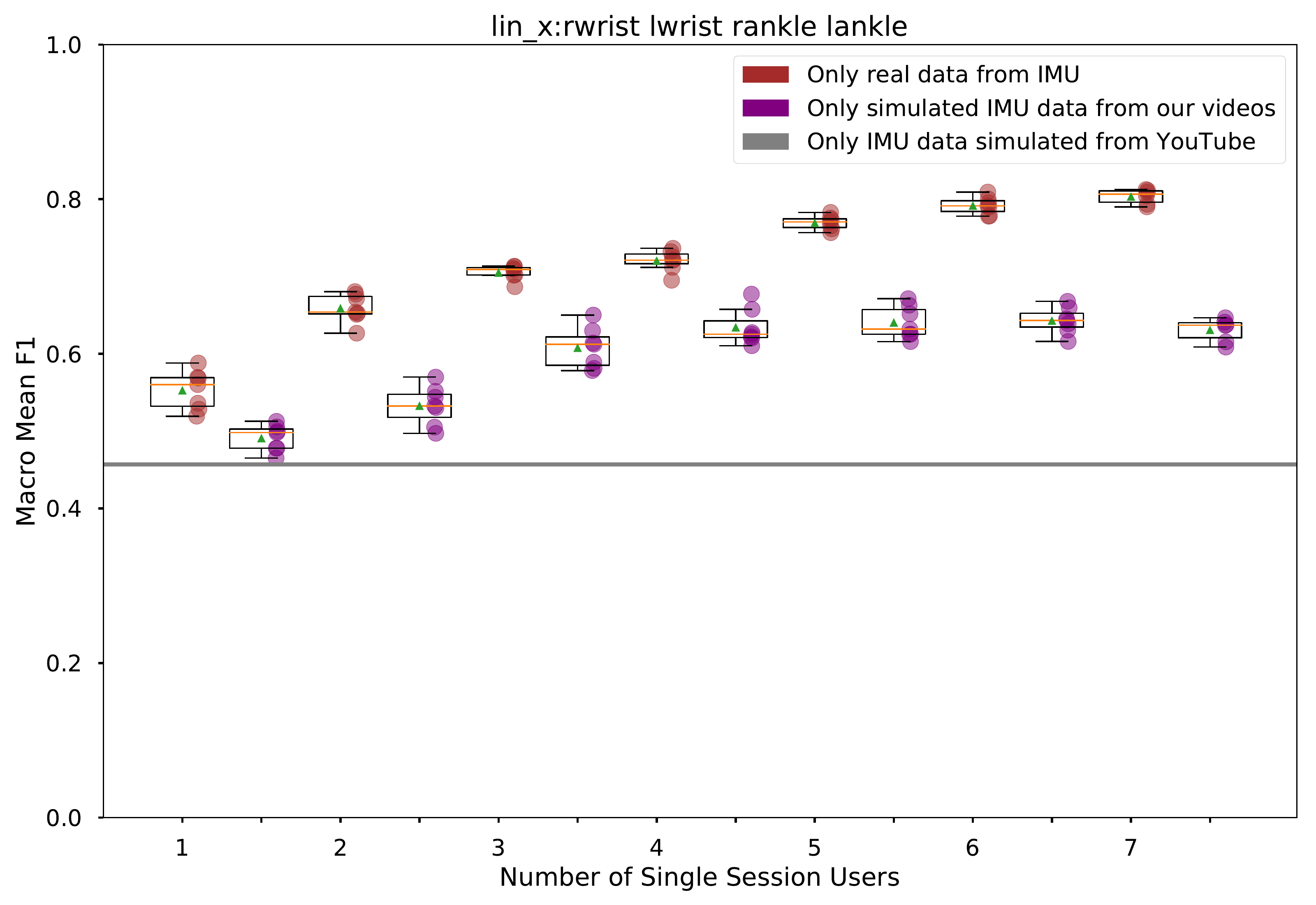}
    \caption{Results when using the linear acceleration in the limb axis. }
    %\label{fig:results_linx_more}
    \label{fig:results_linax}
\end{figure}

%The overall results for the best pre-processing strategy (???) as determined from above are shown in Figures \ref{fig:results_acc}, \ref{fig:results_both}, \ref{fig:results_gyr} and ??? for different sensor combinations and training set compositions.

%Our YouTube only model is very competitive when using accelerometer norm, with results around $0.6$ macro mean F1 score, which is better than a single real user. The simulated data from our videos gets better and better as more users are added, and it alone outperforms real users for $1$ or $2$ real users. From then on, the simulated data alone does not outperform the real one, but adding YouTube simulated data improves model performance.

%\subsubsection{Effect of Different Pre-processing Methods on Classification Results}

%\subsubsection{Signal Level Effect Different Pre-Processing Methods}

%\subsubsection{Mixing Real and Simulated Training Data}

 %\ref{fig:results_acc}, \ref{fig:results_both}, \ref{fig:results_gyr})

%\subsubsection{Different Sensor Sets}
%Compared to our previous work, this time the regression model has never seen the target exercises. Thus, it is very encouraging to see results as in Figure \ref{fig:results_acc}. 
%Regarding gyro norm results, depicted in Figure \ref{fig:results_gyr}, the simulated data by itself is not that good, specially the YouTube data alone. On the other hand, adding this YouTube data is still beneficial with $1$ or $2$ users. Moreover, much of the real performance can be recovered having one or two real users.
%If we combine accelerometer and gyroscope norm, we get results as in Figure \ref{fig:results_both}. In this case we can see patterns similar to using only one modality, that is, adding simulated data improves most for $1$ or $2$ real users.

%
%
\section{Conclusion and Future Work}\label{sec_conclusion}
Overall we can conclude that the proposed approach of generating simulated IMU signals through regression directly from video extracted postures holds promise for the creation of large labeled training datasets for wearable activity recognition without the need to actually record and label data with the sensors. 
In the long term it can contribute to the creation of training datasets that in size are comparable to what exists in computer vision today and thus facilitate a more profound impact of deep learning techniques on wearable sensors based HAR. 

Clearly the work described here is just a first step towards this vision. Based on the results described and discussed in the previous section we consider the following to be the most promising next steps that we will investigate:
\begin{enumerate}
    \item Extending the training data set for our regression model, in terms of the number of users, the variety of motions and the number of sensor placements.
    \item Collecting a large set of online videos for a more diverse set of activities, and further testing and fine-tuning our model on them.  
    \item Extending the regression model to handle more types of sensor signals (e.g. acceleration and gyro on each axis). This includes considering relations between sensor signals e.g.  the orientation of the wrist IMU in the the frame of reference of the hip or back mounted IMU which is often used as relevant information in wearable activity recognition systems.  
    \item Exploring image segmentation based detection of "down" direction and including the angle towards the down vector in the features that we feed our regression model. This will be particularly important as we start looking at more and more complex sensor signals as mentioned above.
    \item Exploring for training the regression  the concept of  Physically Informed Networks\cite{raissi2019physics} that incorporate physical constraints as regularization terms. The idea is to build a regression model that for example  simultaneously train the ax,ay and az component of the acceleration and the corresponding norm $|a|$ and embedding the $|a|=\sqrt{ax^2+ay^2+az^2}$ in the regularization term. Relationships between the accelerometer and the gyroscope sensor signal at the same location and temporal conditions could also be considered. These are all especially relevant for generating more complex sensor signals as described in the previous points
    \item We could also explore applying our method to (more) easily identify locations where sensors should be placed on the body.  Before recording sensor data, videos of the target activities can be collected and sensor values simulated for different on-body locations, which then can help select the best locations to place them when recording training data.
\end{enumerate}

\section{Acknowledgments}\label{sec_ack}
The simulations were executed on the high performance cluster “Elwetritsch” at the TU Kaiserslautern which is part of the “Alliance of High Performance Computing Rheinland-Pfalz” (AHRP). We kindly acknowledge the support under the AI Enhanced Cognition and Learning project, Social Wear and Humane AI Net in general.

%\bibliographystyle{unsrt}  
%\bibliography{references}  %%% Remove comment to use the external .bib file (using bibtex).
%%% and comment out the ``thebibliography'' section.

\end{document}